\documentclass[letterpaper, 10 pt, conference]{ieeeconf}  % Comment this line out if you need a4paper

\IEEEoverridecommandlockouts                              % This command is only needed if 
\usepackage{epsfig} % for postscript graphics files
\usepackage{times} % assumes new font selection scheme installed
\usepackage{amsmath} % assumes amsmath package installed
\usepackage{amssymb}  % assumes amsmath package installed

\usepackage[export]{adjustbox}

\title{\LARGE \bf
RISE-SLAM: A Resource-aware Inverse Schmidt Estimator for SLAM
}

\author{Tong Ke, Kejian J. Wu, and Stergios I. Roumeliotis
\thanks{This work was supported by the National Science Foundation (IIS-1328722, IIS-1637875).}
\thanks{The authors are with the Univ. of Minnesota, Minneapolis, MN, USA. {\tt\small kexxx069|wuxxx834|stergios@umn.edu}}
}
\renewcommand{\baselinestretch}{0.96}
\begin{document}

\maketitle
\thispagestyle{empty}
\pagestyle{empty}

\newtheorem{remark}{Remark}
\newtheorem{proposition}{Proposition}
\newtheorem{corollary}{Corollary}
\newtheorem{lemma}{Lemma}
\def\q{{\bar q}}
\def\mL{{\mathfrak{L}}}
\def\one{{\scriptscriptstyle {1}}}
\def\two{{\scriptscriptstyle {2}}}
\def\three{{\scriptscriptstyle {3}}}
\def\four{{\scriptscriptstyle {4}}}
\def\five{{\scriptscriptstyle {5}}}
\def\T{{\scriptscriptstyle {T}}}
\def\sG{{\scriptscriptstyle {G}}}
\def\sL{{\scriptscriptstyle {L}}}
\def\sZ{{\scriptscriptstyle {Z}}}
\def\sY{{\scriptscriptstyle {Y}}}
\def\sX{{\scriptscriptstyle {X}}}
\def\sYZ{{\scriptscriptstyle {Y\!Z}}}
\def\sf{{\scriptscriptstyle {f}}}
\def\sI{{\scriptscriptstyle {I}}}
\def\sIk{{\scriptscriptstyle {I_{k}}}}
\def\sIi{{\scriptscriptstyle {I_{i}}}}
\def\sIkn{{\scriptscriptstyle {I_{k+n-1}}}}
\def\sIki{{\scriptscriptstyle {I_{ki}}}}
\def\sIkk{{\scriptscriptstyle {I_{k+1}}}}
\def\sIktd{{\scriptscriptstyle {I_{k + t}}}}
\def\sR{{\scriptscriptstyle {R}}}
\def\sRi{{\scriptscriptstyle {R_i}}}
\def\sRj{{\scriptscriptstyle {R_{i+1}}}}
\def\sRm{{\scriptscriptstyle {R_m}}}
\def\sRm1{{\scriptscriptstyle {R_{m+1}}}}
\def\sRjhat{{\scriptscriptstyle {{R}_{i+1}}}}
\def\sRihat{{\scriptscriptstyle {{R}_{i}}}}
\def\sCk{{\scriptscriptstyle {C_k}}}
\def\sCki{{\scriptscriptstyle {C_{ki}}}}
\def\sC{{\scriptscriptstyle {C}}}
\def\sCihat{{\scriptscriptstyle {{C}_i}}}
\def\sCk1{{\scriptscriptstyle {C_{k+1}}}}
\def\sCktd{{\scriptscriptstyle {C_{k + t}}}}
\def\sLp{{\scriptscriptstyle {L^{\prime}}}}
\def\D{\mathbf {Diag}}
\def\qh{\hat {\bar q}}
\def\ph{\mathbf {\hat p}}
\def\p{\mathbf {p}}
\def\krI{\otimes I_{2 \times 2}}
\def\I3{\mathbf I_{3}}
\def\dth{\boldsymbol \delta \boldsymbol\theta}
\def\dthh{\boldsymbol \delta \hat{\boldsymbol\theta}}
\def\kfp{{\mbox{{\footnotesize $(k+1)$}}}}
\newcommand{\skewm}[1]{\ensuremath{\lfloor #1\, \times \rfloor}}
\newcommand{\qd}[2]{\ensuremath{{ }^{#1}\dot{\bar q}_{#2}}}
\newcommand{\qhat}[2]{\ensuremath{{ }^{#1}\hat{\bar q}_{#2}}}
\newcommand{\qhatd}[2]{\ensuremath{{ }^{#1}\dot{\hat{\bar q}}_{#2}}}
\newcommand{\phat}[2]{\ensuremath{{ }^{#1}\hat{ \mathbf p}_{#2}}}
\newcommand{\mh}[2]{\ensuremath{\hat{\mathbf{#1}}_{#2}}}
\newcommand{\C}{{\mathbf C}}
\newcommand{\X}{{\color{red}[X]}}
\newcommand{\bs}[1]{\boldsymbol{#1}}

\newcommand{\leftexp}[2]{{\vphantom{#2}}{}^{#1}{#2}}
\newcommand{\ROBOTvECTOR}[3]{{}^{#1}\boldsymbol{{#2}}_{#3}}
\newcommand{\skewsymm}[1]{\begin{bmatrix}#1{}_{\times}\end{bmatrix}}
\newcommand{\alp}[2]{\textbf{A}_{#1#2}}
\newcommand{\bht}[2]{\textbf{B}_{#1#2}}

\newcommand{\uu}[1]{{\boldsymbol{\Upsilon}}_{#1}}
\newcommand{\ww}[2]{{\boldsymbol{\Psi}}_{#1#2}}
\newcommand{\vv}[1]{{\boldsymbol{\Lambda}}_{#1}}

%%%%%%%%%%%%%%%%%%%%%%%%%%%%%%%%%%%%%%%%%%%%%%%%%%%%%%%%%%%%%%%%%%%%%%%%%%%%%%%%
\begin{abstract}
    In this paper, we present the RISE-SLAM algorithm for performing visual-inertial simultaneous localization and mapping (SLAM), while improving estimation consistency.
Specifically, in order to achieve real-time operation, existing approaches often assume previously-estimated states to be perfectly known, which leads to inconsistent estimates.
Instead, based on the idea of the Schmidt-Kalman filter, which has processing cost linear in the size of the state vector but quadratic memory requirements, we derive a new consistent approximate method in the information domain, which has linear memory requirements and adjustable (constant to linear) processing cost.
In particular, this method, the resource-aware inverse Schmidt estimator (RISE), allows trading estimation accuracy for computational efficiency.
Furthermore, and in order to better address the requirements of a SLAM system during an exploration vs. a relocalization phase, we employ different configurations of RISE (in terms of the number and order of states updated) to maximize accuracy while preserving efficiency.
Lastly, we evaluate the proposed RISE-SLAM algorithm on publicly-available datasets and demonstrate its superiority, both in terms of accuracy and efficiency, as compared to alternative visual-inertial SLAM systems.
%
%We apply the proposed RISE to obtain an online solution to visual-inertial SLAM, with the capability of switching between exploration in unknown areas and relocalization within previously-mapped scenes. As compared to state-of-the-art visual-inertial SLAM systems on public datasets, our implementation of the proposed algorithm achieves superior performance in terms of localization accuracy and processing time, with real-time running speed even on mobile phones.

\end{abstract}

%%%%%%%%%%%%%%%%%%%%%%%%%%%%%%%%%%%%%%%%%%%%%%%%%%%%%%%%%%%%%%%%%%%%%%%%%%%%%%%%
\section{Introduction and Related Work}
Simultaneous localization and mapping (SLAM) is necessary in a wide range of applications, such as robot navigation in GPS-denied areas, autonomous driving, and augmented/virtual reality. Recently, successful vision-only SLAM systems have emerged that employ one or multiple cameras~\cite{klein2007parallel,engel2014lsd,mur2015orb}.
Another popular choice is to combine the visual information with inertial data, from an inertial measurement unit (IMU), for increased robustness and accuracy~\cite{Mourikis07,KFSLAMJ,qin2018vins,mur2017visual,liu2018ice}.
In both cases, it is well known that under certain assumptions, finding the Maximum a Posteriori (MAP) estimate for SLAM can be cast as a nonlinear batch least-squares (BLS) problem, and the optimal solution, for the camera poses and feature positions, can be obtained in either a batch~\cite{Triggs00,Dellaert06squareroot} or an incremental~\cite{isam,isam2} form.
These optimal approaches, however, have an increasing processing cost with time, typically between linear and quadratic in the number of poses and features, and thus cannot provide high-frequency estimates when operating inside large areas.
On the other end of the spectrum, visual(-inertial) odometry systems~\cite{Mourikis07,engel13svo,hesch2014consistency,KFSLAMJ,SQRT_Invf,forster2017svo,bloesch2017iterated,engel18dso} focus their optimization over only a bounded sliding window of recent poses. The latency of these methods is typically very low and does not increase with time, but this comes at the expense of an ever-increasing drift in the pose estimates, due to their inability to process loop-closure measurements and perform global adjustment.
%
%In this work, we focus on the problem of full SLAM, that simultaneously create a global map of the environment in order to enable loop closure and relocalization, for its superiority in terms of accuracy.

In order to achieve accuracy and efficiency at the same time, recent visual(-inertial) SLAM systems aim to combine the advantages of both the optimal (global) and the odometry (local) approaches, by employing a multi-thread scheme~\cite{klein2007parallel,mur2015orb,qin2018vins,liu2018ice}:
A frontend thread estimates the current or several recent poses (as well as a local map) in constant time for real-time performance, while a backend thread optimizes, at a higher cost and lower frequency, over the entire trajectory (using either the optimal BLS~\cite{Dellaert06squareroot} or its approximations~\cite{thrun2006graph,Konolige08frameslam,nerurkarc}), and generates more accurate keyframe pose estimates and global maps for relocalization.
To limit the processing cost, however, all these approaches employ approximations, e.g., keyframes involved in the frontend's relocalization are assumed to be \emph{perfectly known}. Ignoring the corresponding uncertainties of these states and their cross correlations with the current states, however, leads to \emph{inconsistent} estimates.\footnote{As defined in~\cite{Julier1997,Jazwinski1970}, a state estimator is consistent if the estimation errors are zero-mean and have covariance matrix smaller than or equal to the one calculated by the estimator. For the purposes of this work, we focus on the covariance requirement. Note that there exist additional sources of inconsistency, due to linearization errors and local minima (see e.g.,~\cite{Huang2010}). In this work, we focus on the inconsistency caused by the assumption that uncertain quantities such as a map, is perfectly known.}
This means that the estimated covariance is unduly small and does not represent correctly the uncertainty of the current state estimates (i.e., it does not offer a reliable measure of the tracking quality). More importantly, combining these overly optimistic estimates with new measurements later on can further degrade the accuracy of the system, as new, precise measurements are weighted less in favor of the current estimates.
In fact, this problem of inconsistency has been acknowledged in the past, and remedies are often used to alleviate its negative impact on estimation accuracy, e.g., by inflating the assumed covariance of the noise corresponding to the relocalization visual observations~\cite{Mourikis09,lynen2015get}. These heuristics, however, offer no guarantees on the estimation consistency or the system's performance.
%To address this issue, in this work, we present a novel SLAM algorithm, that employs only \emph{consistent} approximations, to \emph{efficiently} compute \emph{consistent} estimates.

On the other hand, the sparse extended information filter (SEIF) of~\cite{thrun04seif,walter07eseif} is a consistent approximate SLAM algorithm, whose cost (between linear and cubic in the map's size) though for recovering the state estimate from the information vector makes it prohibitive for real-time operation. Specifically, although approximations involving early-terminating iterative solvers reduce processing during exploration, the required number of iterations for loop closures makes the cost often larger than that of direct solvers.
%
%Both ours and [b] maintain consistency and sparsity of the information matrix and the Cholesky factor, respectively, but through different means: [b] drops proprioceptive meas/nts while we drop information for older states. Although not easy to assess which of the two approximations is better, our advantage to [a,b] is \emph{real-time} performance.
%
%The major bottleneck of these algorithms is the recovery of the state from the information vector, which has cost at linear to cubic in the size of the map. To address this issue, these systems use iterative solvers terminating early, which are only applicable to exploration (where only few states change) but not to loop closures. In the case of loop closure, the required cost is high and non-constant, hence preventing real-time performances.

At this point, we should note that there exists a consistent approximation in the \emph{filtering} domain: The Schmidt-Kalman filter (SKF)~\cite{schmidt1966application}.
The key idea of the SKF is to update \emph{optimally} only a subset of the states (e.g., recent poses and features) and their corresponding covariance and cross correlation terms, while leaving the rest (e.g., past poses and features) unaltered. By doing so, the computational cost is reduced from quadratic to linear in the (potentially very large) size of unchanged states.
Meanwhile, the uncertainty of the past states is correctly accounted for to guarantee consistent estimates.
The SKF and its variants have been applied to the SLAM problem~\cite{guivant2001optimization,julier2001sparse,nerurkar2011power}, where their major drawback is their high memory requirements: Quadratic in the size of \emph{all} states due to the dense covariance matrix. Thus, the SKF cannot be employed in large-scale SLAM.
On the other hand, it is well-known that the information-domain solutions are more suitable for large-scale SLAM, as the Hessian matrix and its corresponding Cholesky factor are sparse~\cite{Triggs00}.
To leverage this fact,~\cite{dutoit17consistent} adapted the SKF to incorporate a previously-computed sparse Cholesky factor of a given map's Hessian. The approach, however, is a filtering one, and can only be used to perform \emph{map-based localization} given an \emph{offline-built} map, but not SLAM.

Motivated by the potential processing savings of the SKF, as well as the low-memory requirements of the Hessian (or equivalently its Cholesky factorization) representation of the uncertainty, in this work, we seek to derive a Schmidt-type estimator in the \emph{information} domain, that we can apply to the SLAM problem.
To do so, we initially derived the exact equivalent of the SKF in its square-root inverse form, i.e., by maintaining the Cholesky factor of the Hessian, since the corresponding portion of the information factor does not change~\cite{inverseschmidt}.
Surprisingly, unlike the case of the SKF, the exact inverse-form Schmidt estimator does \emph{not} provide any computational savings as compared to the optimal solver~\cite{risetechreport}. Moreover, the involved operations introduce a large number of fill-ins, leading to an almost dense information factor. This eventually makes the system too slow, and hence unsuitable for real-time long-term SLAM.

To overcome these limitations, in this work, we introduce the resource-aware inverse Schmidt estimator (RISE), which is derived as a further approximation of the exact inverse Schmidt estimator~\cite{inverseschmidt}.
The key idea behind RISE is to drop a certain portion of the available information, so that: i)~As in the exact inverse Schmidt, past states as well as their corresponding portion of the information factor remain unaltered, while at the same time, ii)~Recent states are updated only \emph{approximately}, instead of \emph{optimally}, so as to reduce both the processing cost and the factor fill-ins.
Hence, RISE achieves both computational and memory efficiency by keeping the information factor sparse.
Meanwhile, it is a \emph{consistent} approximation to the optimal approach, as it only drops information, instead of assuming any state to be perfectly known.
%
%Sqrt form: Enable efficient implementation using single-precision representation and operations.
%
More importantly, RISE allows trading accuracy for efficiency, by adjusting the size of the window of the states selected to be updated.
%
%Smaller windows provide less accurate estimates at higher rates, while larger windows allow for higher accuracy at the cost of longer processing times.
%
In the extreme case when all states are chosen for update, RISE becomes exactly equivalent to the optimal solver without any information loss.

Furthermore, we employ the proposed RISE algorithm in various configurations to realize an accurate and efficient visual-inertial SLAM system, the RISE-SLAM, which maintains a \emph{consistent} sparse information factor corresponding to all estimated states.
Specifically, our system alternates between two modes, \emph{exploration} and \emph{relocalization}, based on the availability of loop-closure measurements.
In order to balance between accuracy and efficiency, in each mode, RISE is employed with various window sizes and different state orders. % to update only the selected portion of the states and the corresponding square-root factor.
Similarly to most recent SLAM systems, our implementation incorporates two threads running in parallel:
A fast frontend thread for estimating the current poses and features at a high frequency, and a lower-rate backend thread for globally adjusting the past states to achieve high accuracy.
A key difference, however, as compared to existing systems that solve \emph{multiple} optimization problems \emph{independently} in different threads~\cite{mur2015orb,qin2018vins,liu2018ice}, is that \emph{RISE-SLAM always solves a \emph{single} optimization problem, partitioned into two components each assigned to one of the two threads}. This is only possible because of the structure of RISE, whose approximation allows focusing resources on only a subset of states at a time.
As a result, in our system, important global corrections from the backend are \emph{immediately} reflected onto the frontend estimates, hence improving the current tracking accuracy.
%
%Existing systems solve \emph{multiple} optimization problems independently in different threads~\cite{mur2015orb,qin2018vins,liu2018ice}. When a global adjustment finishes in the backend, those corrected estimates will help anchoring the current poses in the frontend, but \emph{only if} these backend-adjusted states are still within the horizon of the frontend and involved as fixed parameters in its current optimization problem. In practice, however, this condition may not always hold, eg, if the platform has already moved to exploring a new area after the time for the backend to finish, so that the current frames involved in the frontend do not share any common feature observations with the previous keyframes from the backend. In this case, the frontend estimates cannot be adjusted from the corrections of the backend, and the tracking quality will suffer from this drift until the next loop closure or global adjustment happens.
%Instead, at any instance, our proposed system always solves a \emph{single} optimization problem. When a global adjustment happens, thanks to the structure of the RISE, this costly problem can be partitioned into two complements to be executed in two threads. This way, on the completion of the backend thread, those global corrections can always be reflected immediately onto the frontend estimates, hence improving the current tracking accuracy.
%
In summary, our main contributions are:
\begin{itemize}
\item We derive the resource-aware inverse Schmidt estimator (RISE), which approximates the exact inverse Schmidt and has adjustable processing cost, while preserving sparsity and ensuring consistency.
%To the best our knowledge, they are the first Schmidt-type estimators in the inverse form.
%
\item We introduce RISE-SLAM, for building 3D maps and relocalizing within previously-mapped areas in a consistent manner with constant cost.
\item We implement RISE-SLAM and assess its performance. As compared to state-of-the-art approaches, our algorithm achieves the best performance in terms of estimation accuracy and processing time.
%Moreover, the proposed system is able to run in real time even on resource-constrained platforms such as mobile phones.
%
\end{itemize}

\section{System Overview} \label{sec:overview}
\begin{figure}[t]
    \centering
    \includegraphics[width=0.45\textwidth]{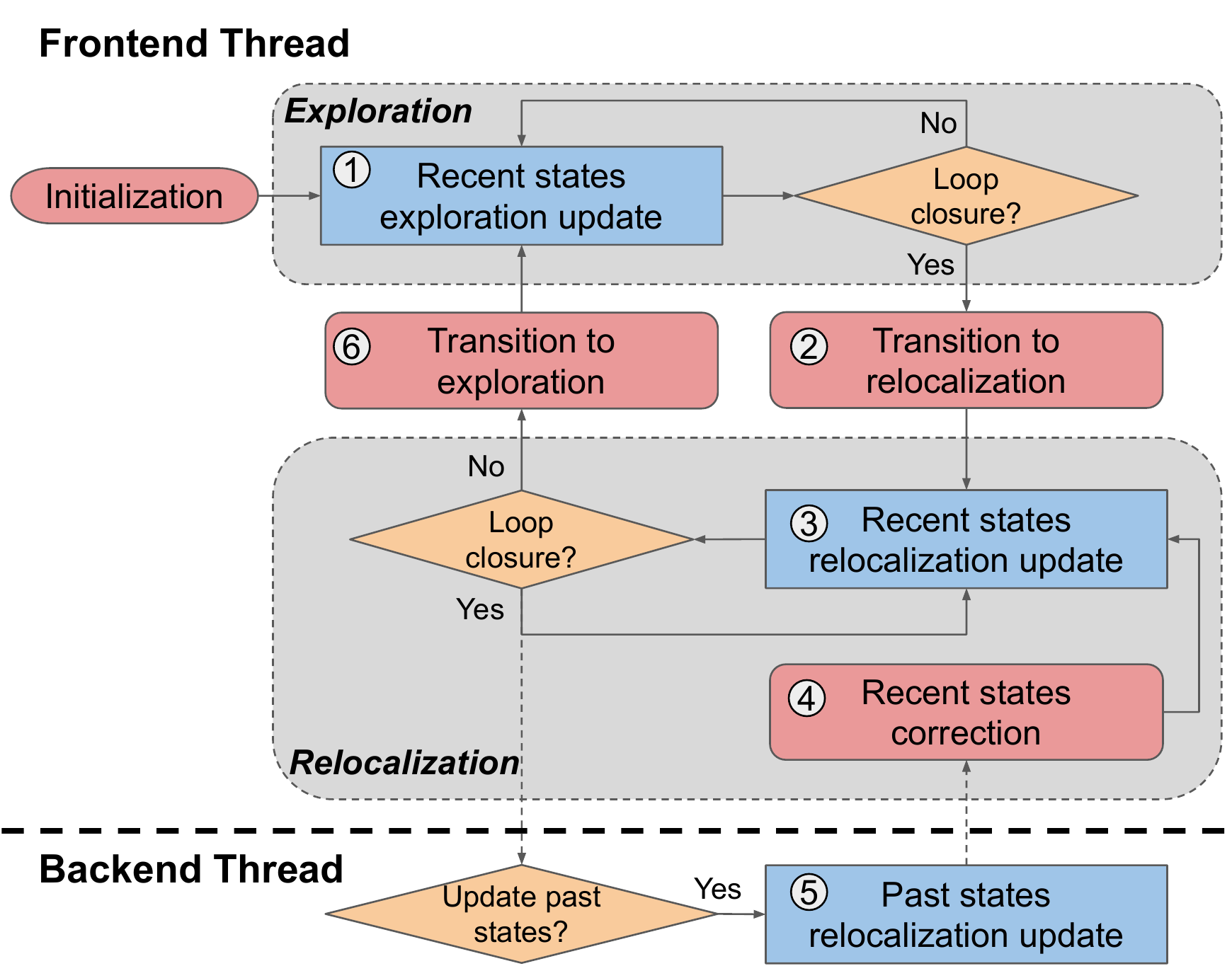}
    \vspace{-1mm}
    \caption{System overview. After initialization, the system starts in exploration mode (Sec.~\ref{ssec:exploration1}), and switches to relocalization mode when a set of loop closures is detected (Sec.~\ref{ssec:torelo}). When in relocalization mode, besides the frontend thread (Sec.~\ref{ssec:frontend}), the system may run a backend thread to perform global adjustment of past states (Sec.~\ref{ssec:backend}), while the frontend thread employs the backend's feedback (i.e., updated trajectory) to correct recent states (Sec.~\ref{ssec:back2front}). Once no loop closures are detected, the system switches back to exploration mode (Sec.~\ref{ssec:toexp}, \ref{ssec:newexp}).}
    \label{fig:overview}
    \vspace{-5mm}
\end{figure}

The proposed visual-inertial SLAM system, whose overview is depicted in Fig.~\ref{fig:overview}, employs an incremental estimator comprising two key modes each addressing the particular needs of the corresponding phases of SLAM: Exploration and Relocalization. During exploration, the IMU-camera pair navigates through a new area. Thus, the feature observations available for processing span only a short window of recent poses.
During relocalization, the IMU-camera pair enters areas it has previously visited, and acquires loop-closure measurements that relate recent camera poses with past ones, thus enabling to remove pose drift.
Such reobservations of past features, however, are typically expensive to process, and for this reason we explicitly distinguish between these two phases, and treat them differently.

Specifically, after initialization, the system begins in exploration mode. The estimator optimizes over a sliding window of recent states involved in the local feature-track measurements (\textcircled{\scriptsize 1} in Fig.~\ref{fig:overview}), with \emph{constant} cost (determined by the window size).
Once loop-closure measurements are detected, the system enters the relocalization mode (\textcircled{\scriptsize 2} in Fig.~\ref{fig:overview}), and spans two threads to process local feature-track measurements as well as loop-closure observations.
In the frontend, the system estimates a sliding window of recent states using \emph{both} types of visual measurements (\textcircled{\scriptsize 3} in Fig.~\ref{fig:overview}) with \emph{constant} cost. 
This is a key novelty of our work and the process for accomplishing this in a consistent manner is detailed in Sec.~\ref{ssec:frontend}.
The optimization of other (past) states (\textcircled{\scriptsize 5} in Fig.~\ref{fig:overview}), which has approximately linear cost in their size, is assigned to the backend.
Note that the two threads run independently so that the frontend can add new states and provide estimates even if the backend is still running.
Once the backend finishes updating the past states, the frontend employs its feedback to correct the recent states (\textcircled{\scriptsize 4} in Fig.~\ref{fig:overview}).
Once all the states are globally adjusted, we only need to run the frontend to update recent states (\textcircled{\scriptsize 3} in Fig.~\ref{fig:overview}).
Though we could enable the backend optimization whenever the backend thread is idle, in order to save processing, in our implementation we choose to run the backend only once during each relocalization phase.
Finally, when there are no more loop-closure measurements available, the system switches back to the exploration mode (\textcircled{\scriptsize 6} in Fig.~\ref{fig:overview}).

In what follows, we introduce the key algorithmic components and contributions necessary for realizing the proposed estimation scheme.
Specifically, we first provide an overview of the optimal square-root inverse estimator (see Sec.~\ref{ssec:optimal}) used during exploration, and then present its Schmidt-based approximation (see Sec.~\ref{ssec:schmidt}) employed later on for reducing the computational cost of relocalization.
%with adjustable Schmidt applied. 
%Unlike VINS-Mono and ICE-BA, which execute another optimization for global mapping than their tracking and rely on later observation of the map to relocalize, our system is a single estimator where local tracking and global mapping are performed in a single optimization problem. The information obtained from global adjustment can be reflected to the current estimated pose immediately. In addition, it is in square-root form, which is more numerically stable and hence allows to use single-precision numbers. Single-precision numbers consume less memory and their operations are usually faster. ARM architecture, used by most modern mobile devices, also features an instruction set NEON, which 

%\textcolor{blue}{scheme of updating a portion of all states, recent and past, need corresponding estimators that enable this scheme. Next section introduce the novel estimators for this.}
\section{Square-root Inverse Estimators for SLAM}\label{sec:estimators}
In this section, we discuss SLAM estimators in the square-root inverse form. We start by denoting the state vector to be estimated as $\mathbf{x}$, which comprises IMU poses and feature positions, and extends with new states as time goes by. At every step, the estimator maintains a prior cost term of the current state estimate, $\left\|\mathbf{R}(\mathbf{x}-\hat{\mathbf{x}})\right\|^2$, 
where $\mathbf{R}$ is the upper-triangular information factor matrix (i.e., the Cholesky factor of the Hessian) and $\hat{\mathbf{x}}$ is the current estimate of $\mathbf{x}$. As new visual or inertial measurements arrive, they contribute another cost term (after linearization), $\left\|\mathbf{H}(\mathbf{x}-\hat{\mathbf{x}})-\mathbf{r}
\right\|^2$, where $\mathbf{H}$ and $\mathbf{r}$ are the measurement Jacobian and residual, respectively.\footnote{We follow~\cite{SQRT_Invf} for the state parameterization, as well as for the visual-inertial measurement processing and cost term formulations.} Then, the updated state estimate, $\hat{\mathbf{x}}^\oplus$, is found by minimizing the cost function:
\begin{align}
    \mathcal{C}&=\left\|\mathbf{R}(\mathbf{x}-\hat{\mathbf{x}})
    \right\|^2 +\left\|\mathbf{H}(\mathbf{x}-\hat{\mathbf{x}})-\mathbf{r}
    \right\|^2\\
    \hat{\mathbf{x}}^\oplus&=\arg\min_{\scriptscriptstyle\mathbf{x}} \mathcal{C} \label{eq:C}
\end{align}
\subsection{Optimal Estimator}\label{ssec:optimal}
The optimal solution of \eqref{eq:C} can be computed as:
\begin{align}
    \mathcal{C}=\left\|\begin{bmatrix}
    \mathbf{R} \\
    \mathbf{H}
    \end{bmatrix}
    \left(\mathbf{x}-\hat{\mathbf{x}}\right)-\begin{bmatrix}\mathbf{0}\\\mathbf{r}\end{bmatrix}
    \right\|^2
    %&=\left\|\mathbf{Q}^\T\begin{bmatrix}
    %\mathbf{R} \\
    %\mathbf{H}
    %\end{bmatrix}
    %\left(\mathbf{x}-\hat{\mathbf{x}}\right)-\mathbf{Q}^\T\begin{bmatrix}\mathbf{0}\\\mathbf{r}\end{bmatrix}
    %\right\|^2\nonumber\\
    &=\left\|\begin{bmatrix}\mathbf{R}^\oplus\\\mathbf{0}\end{bmatrix}
    \left(\mathbf{x}-\hat{\mathbf{x}}\right)-\begin{bmatrix}\mathbf{r}^\oplus\\\mathbf{e}\end{bmatrix}
    \right\|^2 \nonumber \\
    \Rightarrow \hat{\mathbf{x}}^\oplus&=\hat{\mathbf{x}}+{\mathbf{R}^\oplus}^{-1}\mathbf{r}^\oplus
\end{align}
where we have performed the following QR factorization:
\begin{align}
    \begin{bmatrix}
    \mathbf{R} \\
    \mathbf{H}
    \end{bmatrix}=\mathbf{Q}\begin{bmatrix}\mathbf{R}^\oplus\\\mathbf{0}\end{bmatrix},\ 
    \begin{bmatrix}\mathbf{r}^\oplus\\\mathbf{e}\end{bmatrix}\triangleq\mathbf{Q}^\T\begin{bmatrix}\mathbf{0}\\\mathbf{r}\end{bmatrix}
\end{align}

The main advantage of this estimator is its optimality in minimizing the mean square error. Additionally, it is very efficient during the \emph{exploration} phase, if the states in $\mathbf{x}$ follow a \emph{chronological} order. Specially, when only local feature-track measurements are available, as described in~\cite{isam}, the QR factorization needs to involve only the bottom-right part of $\mathbf{R}$, which corresponds to recent states. Thus, the cost remains constant, irrespective of the size of the entire state vector $\mathbf{x}$.
In contrast, during \emph{relocalization}, this estimator becomes very inefficient for processing loop-closure measurements, which involve both recent and past states. In this case, the size of the submatrix of $\mathbf{R}$ involved in the QR factorization increases significantly, making the cost at least linear in the size of $\mathbf{x}$. Since this becomes prohibitively expensive, especially when navigating in large areas, in what follows, we introduce alternative approximate estimators that reduce the computational cost, while preserving consistency.
\subsection{Estimators based on the Schmidt Approximation}\label{ssec:schmidt}
As mentioned earlier, the Schmidt approximation, which was introduced originally for the Kalman filter~\cite{schmidt1966application}, is consistent. The key idea behind it is to save processing cost by updating only a subset of the states while leaving the rest unaltered. In what follows, we discuss its equivalent in the square-root inverse form, identify its limitations, and introduce RISE, an efficient alternative to it.
We start by partitioning the state vector $\mathbf{x}$ into two parts: $\mathbf{x}_\one$ and $\mathbf{x}_\two$. %, and divide $\mathbf{R}$ and $\mathbf{H}$ accordingly. 
Now the prior term can be written as
\begin{align}
    \left\|\mathbf{R}(\mathbf{x}-\hat{\mathbf{x}})
    \right\|^2=\left\|\begin{bmatrix}
    \mathbf{R}_{\scriptscriptstyle11} & \mathbf{R}_{\scriptscriptstyle12} \\
    & \mathbf{R}_{\scriptscriptstyle22}
    \end{bmatrix}
    \begin{bmatrix}\mathbf{x}_\one-\hat{\mathbf{x}}_\one\\
    \mathbf{x}_\two-\hat{\mathbf{x}}_\two\end{bmatrix}\right\|^2 \label{eq:prior}
\end{align}
Employing the idea of Schmidt, we use the measurements to update the estimate of $\mathbf{x}_\one$ to $\hat{\mathbf{x}}_\one^\oplus$ but keep $\hat{\mathbf{x}}_\two$ the same.
By doing so, the posterior term should become
\begin{align}
    \left\|\begin{bmatrix}
    \mathbf{R}_{\scriptscriptstyle11}^\oplus & \mathbf{R}_{\scriptscriptstyle12}^\oplus \\
    & \mathbf{R}_{\scriptscriptstyle22}
    \end{bmatrix}
    \begin{bmatrix}\mathbf{x}_\one-\hat{\mathbf{x}}_\one^\oplus\\
    \mathbf{x}_\two-\hat{\mathbf{x}}_\two\end{bmatrix}
    \right\|^2
\end{align}

A property of the Schmidt approximation in this square-root inverse form is that $\mathbf{R}_{\scriptscriptstyle22}$ remains the same, which does not hold if we change the state order (update $\mathbf{x}_\two$ but not $\mathbf{x}_\one$). For this reason, it is preferable to put the states to be updated on the upper part of $\mathbf{x}$. In practice, we typically focus more on recent states than past states, so the states must be organized in \emph{reverse chronological} order in order to apply the Schmidt approximation. This is in stark contrast to the preferred state order for the case of the optimal estimator during exploration. The ramifications of this order switching will become evident when we discuss them in Sec.~\ref{sec:exploration}-\ref{sec:relocalization}.

Among all Schmidt estimators, the exact Schmidt~\cite{schmidt1966application} yields the optimal solution for $\mathbf{x}_\one$. Surprisingly (and quite unfortunately), its exact equivalent in the inverse form, which we derived in~\cite{inverseschmidt}, called the inverse Schmidt estimator (ISE), has no speed advantage over the optimal estimator, and causes fill-ins when applied to SLAM. For this reason, in this work we introduce an approximation to ISE, which has lower cost, the \emph{resource-aware inverse Schmidt estimator} (RISE). This can be derived following the outline in~\cite{inverseschmidt} by setting $\mathbf{R}_{\scriptscriptstyle22}$ to infinity in the steps of ISE, and is summarized hereafter. First, we rewrite the cost function in~\eqref{eq:C} as
\begin{align}
    \mathcal{C}%&=\left\|\begin{bmatrix}
    %\mathbf{R}_{\scriptscriptstyle11} & \mathbf{R}_{\scriptscriptstyle12} \\
    %& \mathbf{R}_{\scriptscriptstyle22}
    %\end{bmatrix}
    %\begin{bmatrix}\mathbf{x}_\one-\hat{\mathbf{x}}_\one\\
    %\mathbf{x}_\two-\hat{\mathbf{x}}_\two\end{bmatrix}
    %\right\|^2\nonumber\\&+\left\|\begin{bmatrix}\mathbf{H}_\one & \mathbf{H}_\two\end{bmatrix}\begin{bmatrix}\mathbf{x}_\one-\hat{\mathbf{x}}_\one\\
    %\mathbf{x}_\two-\hat{\mathbf{x}}_\two\end{bmatrix}-\mathbf{r}
    %\right\|^2\nonumber\\
    &=\left\|\begin{bmatrix}\mathbf{R}_{\scriptscriptstyle11} & \mathbf{R}_{\scriptscriptstyle12} \\
    \mathbf{H}_\one & \mathbf{H}_\two\end{bmatrix}
    \begin{bmatrix}\mathbf{x}_\one-\hat{\mathbf{x}}_\one\\
    \mathbf{x}_\two-\hat{\mathbf{x}}_\two\end{bmatrix}-\begin{bmatrix}\mathbf{0}\\\mathbf{r}\end{bmatrix}
    \right\|^2+\left\|\mathbf{R}_{\scriptscriptstyle22}(\mathbf{x}_\two-\hat{\mathbf{x}}_\two)
    \right\|^2\nonumber\\
    % &=\left\|\begin{bmatrix}\mathbf{R}_{\scriptscriptstyle11}^\oplus & \mathbf{R}_{\scriptscriptstyle12}^\oplus \\
    % & \mathbf{H}_\two^\oplus\end{bmatrix}
    % \begin{bmatrix}\mathbf{x}_\one-\hat{\mathbf{x}}_\one\\
    % \mathbf{x}_\two-\hat{\mathbf{x}}_\two\end{bmatrix}-\begin{bmatrix}\mathbf{r}_\one^\oplus\\\mathbf{e}_\one\end{bmatrix}
    % \right\|^2+\left\|\mathbf{R}_{\scriptscriptstyle22}(\mathbf{x}_\two-\hat{\mathbf{x}}_\two)
    % \right\|^2
    &=\left\|\begin{bmatrix}\mathbf{R}_{\scriptscriptstyle11}^\oplus & \mathbf{R}_{\scriptscriptstyle12}^\oplus \\
    & \mathbf{H}_\two^\oplus\\
    & \mathbf{R}_{\scriptscriptstyle22}\end{bmatrix}
    \begin{bmatrix}\mathbf{x}_\one-\hat{\mathbf{x}}_\one\\
    \mathbf{x}_\two-\hat{\mathbf{x}}_\two\end{bmatrix}-\begin{bmatrix}\mathbf{r}_\one^\oplus\\\mathbf{e}_\one\\\mathbf{0}\end{bmatrix}
    \right\|^2\label{eq:qr1}
    %\nonumber\\
    %&=\left\|\begin{bmatrix}
    %\mathbf{R}_{\scriptscriptstyle11}^\oplus & \mathbf{R}_{\scriptscriptstyle12}^\oplus\\
    % & \mathbf{R}_{\scriptscriptstyle22}\\
    % & \mathbf{H}_\two^\oplus
    %\end{bmatrix}
    %\begin{bmatrix}\mathbf{x}_\one-\hat{\mathbf{x}}_\one\\
    %\mathbf{x}_\two-\hat{\mathbf{x}}_\two\end{bmatrix}-\begin{bmatrix}\mathbf{r}_\one^\oplus\\\mathbf{0}\\\mathbf{e}_\one\end{bmatrix}
    %\right\|^2
\end{align}
where the following QR factorization was performed:
\begin{align}
    \begin{bmatrix}
    \mathbf{R}_{\scriptscriptstyle11} \\
    \mathbf{H}_\one
    \end{bmatrix}=\mathbf{Q}_\one\begin{bmatrix}\mathbf{R}_{\scriptscriptstyle11}^\oplus\\\mathbf{0}\end{bmatrix}\label{eq:qr1_def}
\end{align}
and
\begin{align}
    \begin{bmatrix}\mathbf{R}_{\scriptscriptstyle12}^\oplus\\\mathbf{H}_\two^\oplus\end{bmatrix}\triangleq\mathbf{Q}_\one^\T\begin{bmatrix}\mathbf{R}_{\scriptscriptstyle12}\\\mathbf{H}_\two\end{bmatrix},\ 
    \begin{bmatrix}\mathbf{r}_\one^\oplus\\\mathbf{e}_\one\end{bmatrix}\triangleq\mathbf{Q}_\one^\T\begin{bmatrix}\mathbf{0}\\\mathbf{r}\end{bmatrix}\label{eq:qr1_rhs}
\end{align}
% \begin{figure}[ht]
%     \centering
%     \includegraphics[width=0.45\textwidth]{fig/est1.pdf}
%     \caption{Structure of the information factor when applying the optimal estimator. $\mathbf{R}$ has a banded diagonal in SLAM problems. $\mathbf{H}$ has two dense blocks far away from each other, which is typically the case when loop-closure measurements relate recent poses with past ones.}
%     \label{fig:est1}
% \end{figure}
\begin{figure}
    \centering
    \includegraphics[width=0.45\textwidth]{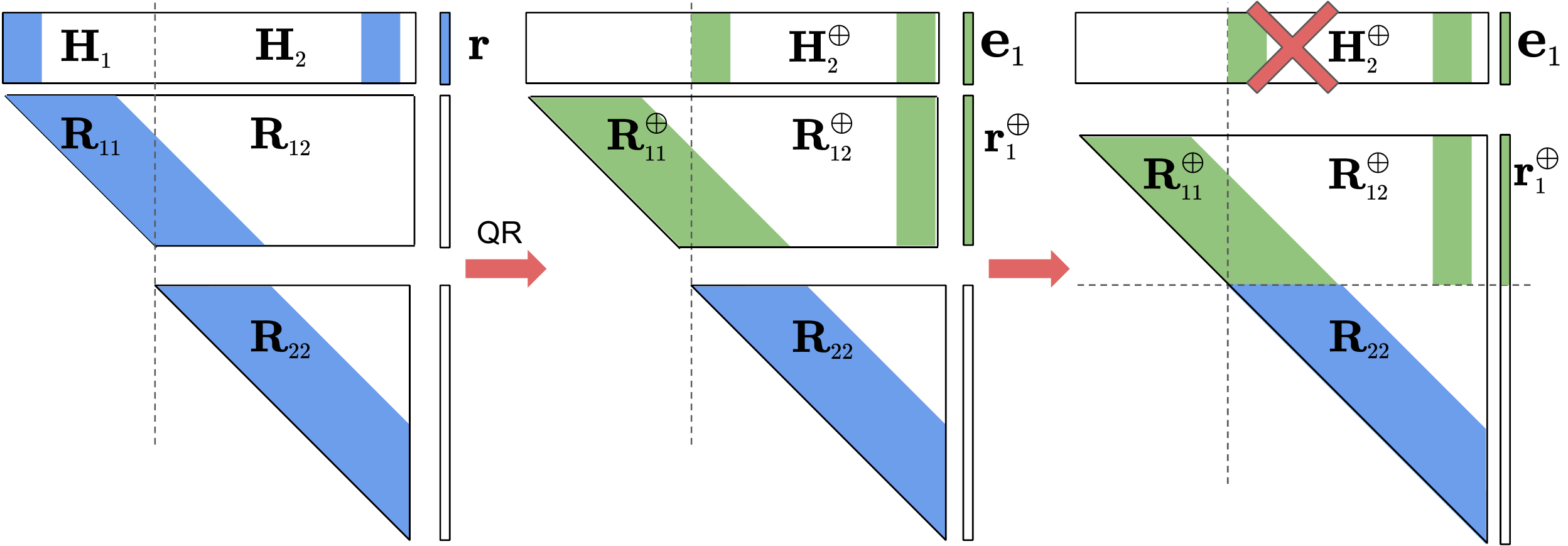}
    \vspace{-1mm}
    \caption{Structure of the information factor when applying RISE. The QR factorization does not involve $\mathbf{R}_{\scriptscriptstyle22}$. We drop the cost term $\left\|\mathbf{H}_\two^\oplus
    \left(\mathbf{x}_\two-\hat{\mathbf{x}}_\two\right)-\mathbf{e}_\one
    \right\|^2$, and combine $\mathbf{R}_{\scriptscriptstyle22}$ with $\mathbf{R}_{\scriptscriptstyle11}^\oplus$ and $\mathbf{R}_{\scriptscriptstyle12}^\oplus$ to form the new cost function $\bar{\mathcal{C}}$ for updating $\mathbf{x}_\one$, while $\mathbf{x}_\two$ remains unchanged. Dropping this cost term is the key approximation of RISE. And by ignoring the information in it, we preserve the sparsity of the Cholesky factor.}
    \label{fig:est2}
    \vspace{-5mm}
\end{figure}
Next, instead of minimizing $\mathcal{C}$, we drop the cost term $\mathbf{x}_\two$, $\left\|\mathbf{H}_\two^\oplus
    \left(\mathbf{x}_\two-\hat{\mathbf{x}}_\two\right)-\mathbf{e}_\one
    \right\|^2$ in~\eqref{eq:qr1}, and minimize (see Fig.~\ref{fig:est2}):
\begin{align}
    \bar{\mathcal{C}}=\left\|\begin{bmatrix}
    \mathbf{R}_{\scriptscriptstyle11}^\oplus & \mathbf{R}_{\scriptscriptstyle12}^\oplus\\
     & \mathbf{R}_{\scriptscriptstyle22}
    \end{bmatrix}
    \begin{bmatrix}\mathbf{x}_\one-\hat{\mathbf{x}}_\one\\
    \mathbf{x}_\two-\hat{\mathbf{x}}_\two\end{bmatrix}-\begin{bmatrix}\mathbf{r}_\one^\oplus\\\mathbf{0}\end{bmatrix}
    \right\|^2 \label{eq:Cbar}
\end{align}
Finally, we update the estimate of $\mathbf{x}_\one$ by setting
\begin{align}
    %\bar{\mathcal{C}}&=\left\|\begin{bmatrix}
    %\mathbf{R}_{\scriptscriptstyle11}^\oplus & \mathbf{R}_{\scriptscriptstyle12}^\oplus\\
    % & \mathbf{R}_{\scriptscriptstyle22}
    %\end{bmatrix}
    %\begin{bmatrix}\mathbf{x}_\one-\mathbf{R}_{\scriptscriptstyle11}^{\oplus-1}\mathbf{r}_\one^\oplus\\\mathbf{x}_\two-\hat{\mathbf{x}}_\two\end{bmatrix}
    %\right\|^2\\
    %\Rightarrow
    \hat{\mathbf{x}}_\one^\oplus&=\arg\min_{\mathbf{x}_\one} \bar{\mathcal{C}}=\hat{\mathbf{x}}_\one+{\mathbf{R}_{\scriptscriptstyle11}^\oplus}^{-1}\mathbf{r}_\one^\oplus
\end{align}
while $\hat{\mathbf{x}}_\two$ remains unchanged.
%
%This simple, yet powerful idea, at the 
As a Schmidt estimator, RISE is \emph{consistent} since it does \emph{not} assume any state as perfectly known. Instead, it only drops information (the term $\left\|\mathbf{H}_\two^\oplus
    \left(\mathbf{x}_\two-\hat{\mathbf{x}}_\two\right)-\mathbf{e}_\one
    \right\|^2$), and correctly updates the cross term $\mathbf{R}_{\scriptscriptstyle12}$ between $\mathbf{x}_\one$ and $\mathbf{x}_\two$. As compared to ISE, RISE computes an approximate estimate for $\mathbf{x}_\one$, whose accuracy loss is negligible when the estimate of $\mathbf{x}_\two$ is precise. In the extreme case, when the uncertainty of $\mathbf{x}_\two$ goes to zero, RISE results in the optimal solution for $\mathbf{x}_\one$ as ISE. For this reason, in practice, we set $\mathbf{x}_\two$ to be states with low uncertainty.
    On the other hand, RISE is significantly more efficient than ISE since the cost of the QR factorization is cubic in the size of $\mathbf{x}_\one$ (instead of $\mathbf{x}$), and introduces no extra fill-ins, thus keeping $\mathbf{R}$ sparse (see the accompanying video for this sparsity in a SLAM simulation).
    If we select $\mathbf{x}_\one$ to contain a small number of states (e.g., a window of recent camera poses and features), the cost is O(1) in the size of $\mathbf{x}$. Although the column size of $\mathbf{R}_{\scriptscriptstyle12}^\oplus$ is the same as the size of $\mathbf{x}_\two$ and can be comparable to that of $\mathbf{x}$, it is sparse in the context of SLAM. As a result, computing $\mathbf{R}_{\scriptscriptstyle12}^\oplus$ is also O(1). 

A key advantage of RISE is that it can trade accuracy for speed by adjusting the size of $\mathbf{x}_\one$. Specifically, during relocalization, we can set $\mathbf{x}_\one=\mathbf{x}$ to obtain an accurate global adjustment if it is the first loop-closure event, or we can use RISE with a small-size $\mathbf{x}_\one$ for an approximate but efficient solution. Furthermore, this global adjustment can be split into two steps, where we first employ RISE with a small sized $\mathbf{x}_\one$, and then we optimize over $\mathbf{x}_\two$, which can run independently in the backend. Through this process, detailed in Sec.~\ref{sec:relocalization}, the frontend maintains its real-time localization capability, while the backend allows taking advantage of loop-closure events after long periods of exploration.

Note also that the optimal estimator for exploration in Sec.~\ref{ssec:optimal} can be considered as a special case of RISE, with $\mathbf{x}_\one=\mathbf{x}$ and a chronological state order. As it will become evident hereafter, by applying RISE, with different state orders/sizes of $\mathbf{x}_\one$, in the two phases of SLAM, we can achieve real-time performance while maintaining consistency.

\section{RISE-SLAM: Exploration}\label{sec:exploration}
In this section, we describe how RISE-SLAM processes local feature tracks (see \textcircled{\scriptsize 1} in Fig.~\ref{fig:overview}) during exploration. We start with the first exploration (Sec.~\ref{ssec:exploration1}), and then discuss the general case, where the system has just switched back to exploration from relocalization (Sec.~\ref{ssec:toexp} and~\ref{ssec:newexp}).
%%%%%%%%%%%%%%%%%%%%%%%%%%%%%%%%%%%%%%
\subsection{First Exploration}\label{ssec:exploration1}
\begin{figure}
    \center
    \includegraphics[width=0.48\textwidth]{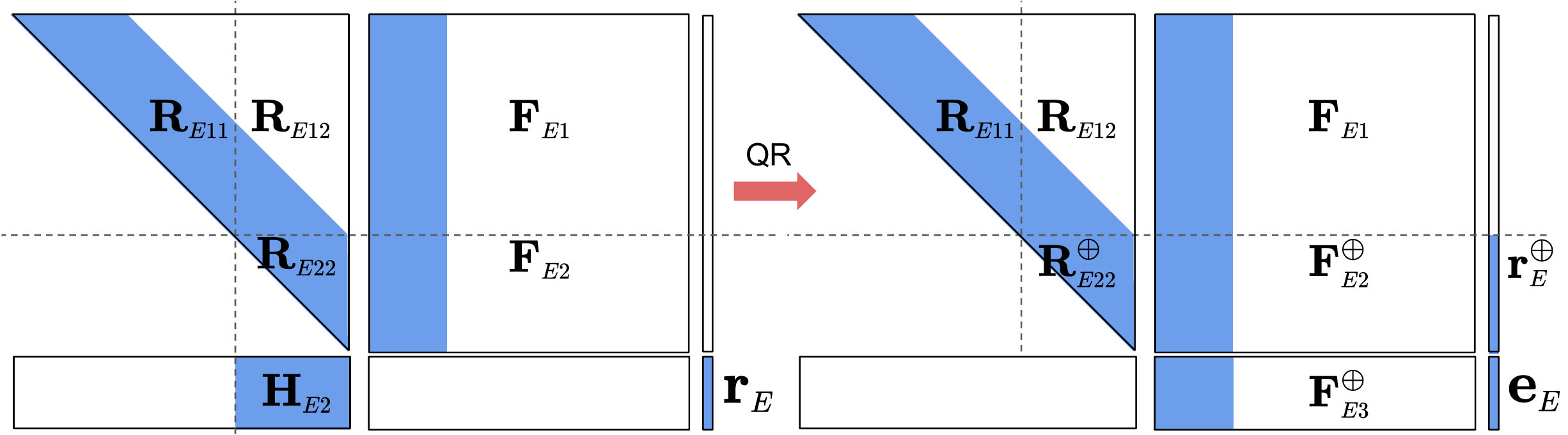}
    \caption{Structure of exploration cost terms before and after an update. $\mathbf{F}_{\scriptscriptstyle E}$ does not exist in Sec.~\ref{ssec:exploration1}, while in Sec.~\ref{ssec:newexp}, it contains the cross information between the new and the old map, where the dense columns on the left correspond to some last states of the old map.}
    \label{fig:exp_new1}
    \vspace{-5mm}
\end{figure}

During the first exploration, we realize the efficiency of the optimal estimator (Sec.~\ref{ssec:optimal}) by organizing the states in \emph{chronological} order~\cite{isam}.
Moreover, we apply RISE with $\mathbf{x}_\one=\mathbf{x}$, which is equivalent to the optimal estimator. Denote the state vector as $\begin{bmatrix}\mathbf{x}_{\scriptscriptstyle E1}^\T & \mathbf{x}_{\scriptscriptstyle E2}^\T\end{bmatrix}^\T$,
where $\mathbf{x}_{\scriptscriptstyle E2}$ comprises a sliding window of \emph{recent} states involved in the local feature-track measurements, while $\mathbf{x}_{\scriptscriptstyle E1}$ contains all other (previous) states.
The cost function to minimize is (see Fig.~\ref{fig:exp_new1})
\begin{align}
    \mathcal{C}_{\scriptscriptstyle E}&=\left\|\begin{bmatrix}
    \mathbf{R}_{\scriptscriptstyle E11} & \mathbf{R}_{\scriptscriptstyle E12} \\
    & \mathbf{R}_{\scriptscriptstyle E22}
    \end{bmatrix}\begin{bmatrix}\mathbf{x}_{\scriptscriptstyle E1}-\hat{\mathbf{x}}_{\scriptscriptstyle E1}\\\mathbf{x}_{\scriptscriptstyle E2}-\hat{\mathbf{x}}_{\scriptscriptstyle E2}\end{bmatrix}
    \right\|^2\nonumber\\&\ \ \ +\left\|\mathbf{H}_{\scriptscriptstyle E2}(\mathbf{x}_{\scriptscriptstyle E2}-\hat{\mathbf{x}}_{\scriptscriptstyle E2})-\mathbf{r}_{\scriptscriptstyle E}
    \right\|^2\label{eq:cexp0}
\end{align}
where the first term is the prior from the previous time step, while the second term corresponds to the new (IMU or feature) measurements. Then, the optimal solution is
    \begin{align}
    \hat{\mathbf{x}}_{\scriptscriptstyle E2}^\oplus&=\hat{\mathbf{x}}_{\scriptscriptstyle E2}+{\mathbf{R}_{\scriptscriptstyle E22}^\oplus}^{-1}\mathbf{r}_{\scriptscriptstyle E}^\oplus\label{eq:x2e}\\
    \hat{\mathbf{x}}_{\scriptscriptstyle E1}^\oplus&=\hat{\mathbf{x}}_{\scriptscriptstyle E1} - \mathbf{R}_{\scriptscriptstyle E11}^{-1}\mathbf{R}_{\scriptscriptstyle E12}{\mathbf{R}_{\scriptscriptstyle E22}^\oplus}^{-1}\mathbf{r}_{\scriptscriptstyle E}^\oplus
\end{align}
which results from rewriting \eqref{eq:cexp0} as
\begin{align}
    \mathcal{C}_{\scriptscriptstyle E}%&=\left\|\begin{bmatrix}
    %\mathbf{R}_{\scriptscriptstyle E11} & \mathbf{R}_{\scriptscriptstyle E12}
    %\end{bmatrix}
    %\left(\mathbf{x}_{\scriptscriptstyle E}-\hat{\mathbf{x}}_{\scriptscriptstyle E}\right)\right\|^2\nonumber\\&+
    %\left\|\begin{bmatrix}
    %\mathbf{R}_{\scriptscriptstyle E22} \\
    %\mathbf{H}_{\scriptscriptstyle E2}
    %\end{bmatrix}
    %\left(\mathbf{x}_{\scriptscriptstyle E2}-\hat{\mathbf{x}}_{\scriptscriptstyle E2}\right)-\begin{bmatrix}\mathbf{0}\\\mathbf{r}_{\scriptscriptstyle E}\end{bmatrix}
    %\right\|^2\nonumber\\
    %&=\left\|\begin{bmatrix}
    %\mathbf{R}_{\scriptscriptstyle E11} & \mathbf{R}_{\scriptscriptstyle E12}
    %\end{bmatrix}
    %\left(\mathbf{x}_{\scriptscriptstyle E}-\hat{\mathbf{x}}_{\scriptscriptstyle E}\right)\right\|^2\nonumber\\&+
    %\left\|\begin{bmatrix}\mathbf{R}_{\scriptscriptstyle E22}^\oplus\\\mathbf{0}\end{bmatrix}
    %\left(\mathbf{x}_{\scriptscriptstyle E2}-\hat{\mathbf{x}}_{\scriptscriptstyle E2}\right)-\begin{bmatrix}\mathbf{r}_{\scriptscriptstyle E}^\oplus\\\mathbf{e}_{\scriptscriptstyle E}\end{bmatrix}
    %\right\|^2\nonumber\\
    %&=\left\|\begin{bmatrix}
    %\mathbf{R}_{\scriptscriptstyle E11} & \mathbf{R}_{\scriptscriptstyle E12} \\
    %& \mathbf{R}_{\scriptscriptstyle E22}^\oplus
    %\end{bmatrix}
    %\left(\mathbf{x}_{\scriptscriptstyle E}-\hat{\mathbf{x}}_{\scriptscriptstyle E}\right)-\begin{bmatrix}\mathbf{0}\\\mathbf{r}_{\scriptscriptstyle E}^\oplus\end{bmatrix}
    %\right\|^2
    %+\left\|\mathbf{e}_{\scriptscriptstyle E}\right\|^2\\
    &=\left\|\begin{bmatrix}
    \mathbf{R}_{\scriptscriptstyle E11} & \mathbf{R}_{\scriptscriptstyle E12} \\
    & \mathbf{R}_{\scriptscriptstyle E22}^\oplus
    \end{bmatrix}
    \begin{bmatrix}\mathbf{x}_{\scriptscriptstyle E1}-\hat{\mathbf{x}}_{\scriptscriptstyle E1}^\oplus\\\mathbf{x}_{\scriptscriptstyle E2}-\hat{\mathbf{x}}_{\scriptscriptstyle E2}^\oplus\end{bmatrix}
    \right\|^2
    +\left\|\mathbf{e}_{\scriptscriptstyle E}\right\|^2\label{eq:cexp0end}
    \end{align}
where $\mathbf{R}_{\scriptscriptstyle E22}^\oplus$ is computed by the following QR factorization:
\begin{equation}
    \begin{bmatrix}
    \mathbf{R}_{\scriptscriptstyle E22} \\
    \mathbf{H}_{\scriptscriptstyle E2}
    \end{bmatrix}=\mathbf{Q}_{\scriptscriptstyle E}\begin{bmatrix}\mathbf{R}_{\scriptscriptstyle E22}^\oplus\\\mathbf{0}\end{bmatrix},\ 
   \begin{bmatrix}\mathbf{r}_{\scriptscriptstyle E}^\oplus\\\mathbf{e}_{\scriptscriptstyle E}\end{bmatrix}=\mathbf{Q}_{\scriptscriptstyle E}^\T\begin{bmatrix}\mathbf{0}\\\mathbf{r}_{\scriptscriptstyle E}\end{bmatrix}\label{eq:qr_first_exploration}
\end{equation}
The impact of this QR factorization on the terms appearing in the cost functions in \eqref{eq:cexp0}-\eqref{eq:cexp0end} is depicted in Fig.~\ref{fig:exp_new1}. These steps are actually analogous to those described in~\cite{isam}, and similarly, the computational complexity is \emph{constant} and only depends on the size of $\mathbf{x}_{\scriptscriptstyle E2}$. As compared to~\cite{isam}, we improve speed during exploration by limiting both the number of features processed at every step and the feature tracks' length (since longer tracks offer more accuracy yet diminishing returns, while increasing the processing cost cubically) based on a preselected size of $\mathbf{x}_{\scriptscriptstyle E2}$. %\textcolor{blue}{Also, \cite{isam} keeps all the states from the beginning while we marginalize some of the IMU states (biases and velocity) to reduce state vector size and improve speed (see~\cite{risetechreport}).}
%Note that $\begin{bmatrix}\mathbf{R}_{\scriptscriptstyle E11} & \mathbf{R}_{\scriptscriptstyle E12}\end{bmatrix}$ remains the same while $\mathbf{R}_{\scriptscriptstyle E22}$ is changed. The cost of computing $\mathbf{R}_{\scriptscriptstyle E22}^\oplus$ and $\hat{\mathbf{x}}_{\scriptscriptstyle E2}^\oplus$ is cubic in terms of the size of $\mathbf{x}_{\scriptscriptstyle E2}$, and constant in terms of the size of $\mathbf{x}$, which makes the update during exploration efficient. To further improve the speed, we have a budget in terms of the number of features and the length of a feature to limit the number of states involved in local measurements.%In addition, we also marginalize some of IMU parameter states (biases and velocities) to reduce the size of $\mathbf{x}_{\scriptscriptstyle E2}$.

%As mentioned in~\cite{isam}, even though back solving for $\hat{\mathbf{x}}_o^\oplus$ takes O(N) time in theory, it is efficient in practice since back solve is much faster than QR factorization. One can also just keep $\mathbf{r}_{\scriptscriptstyle E}^\oplus$ and only perform back substitution when necessary to make the cost constant at every step. Overall, the cost during exploration is constant.
%%%%%%%%%%%%%%%%%%%%%%%%%%%%%%%%%%%%%%%%%%%%%%%%%%%%%%%%%%%%%%%%%%%%%%%%%%%%%%
\subsection{Transition from Relocalization to Exploration} \label{ssec:toexp}
Consider the case when the system is in the relocalization mode (Sec.~\ref{sec:relocalization}) and is about to switch to the exploration mode (i.e., it receives no more loop-closure measurements) (see \textcircled{\scriptsize 6} in Fig.~\ref{fig:overview}).
Due to the opposite state orderings used in these two modes respectively (Sec.~\ref{ssec:schmidt}), we first need to change the order of the recent states from \emph{reverse chronological}, as required in relocalization (Sec.~\ref{sec:relocalization}), to \emph{chronological}, as for exploration (Sec.~\ref{ssec:exploration1}). Specifically, we divide the state vector in relocalization into two parts: $\begin{bmatrix}\mathbf{x}_{\scriptscriptstyle N}^{\prime\T} & \mathbf{x}_{\scriptscriptstyle M}^{\prime\T}\end{bmatrix}^\T$ (the superscript ${}^\prime$ denotes reverse chronological order), where $\mathbf{x}_{\scriptscriptstyle N}^\prime$ contains the \emph{recent} states where no loop-closure measurements are received. Then, we change the state order of $\mathbf{x}_{\scriptscriptstyle N}^\prime$ to chronological, by $\mathbf{x}_{\scriptscriptstyle N}\triangleq\mathbf{P}_{\scriptscriptstyle N}\mathbf{x}_{\scriptscriptstyle N}^\prime$ ($\mathbf{P}_{\scriptscriptstyle N}$ is a permutation matrix). Subsequently, we perform a QR factorization to make the permuted Cholesky factor  upper-triangular again, which is of \emph{constant} cost (see~\cite{risetechreport} for details) regardless of the size of $\mathbf{x}_{\scriptscriptstyle M}^\prime$. After these operations, the cost function has the form
\begin{align}
\mathcal{C}_{\scriptscriptstyle N}
    &=\left\|\begin{bmatrix}
    \mathbf{R}_{\scriptscriptstyle N11} & \mathbf{R}_{\scriptscriptstyle N12}\\
    \end{bmatrix}
    \begin{bmatrix}\mathbf{x}_{\scriptscriptstyle N}-\hat{\mathbf{x}}_{\scriptscriptstyle N}\\\mathbf{x}_{\scriptscriptstyle M}^{\prime}-\hat{\mathbf{x}}_{\scriptscriptstyle M}^{\prime}\end{bmatrix}
    \right\|^2+\mathcal{C}_{\scriptscriptstyle M}\label{eq:cn}
\end{align}
where $\mathcal{C}_{\scriptscriptstyle M}=\left\|\mathbf{R}_{\scriptscriptstyle M}
\left(\mathbf{x}_{\scriptscriptstyle M}^\prime-\hat{\mathbf{x}}_{\scriptscriptstyle M}^\prime\right)
\right\|^2$. Note that the states in $\mathbf{x}_{\scriptscriptstyle M}^\prime$ are considered as an old map, which we do not update during the new exploration, so $\mathcal{C}_{\scriptscriptstyle M}$ will not be used in Sec.~\ref{ssec:newexp}.

After this transition step, a new map (comprising new camera poses and features) begins with $\mathbf{x}_{\scriptscriptstyle N}$, while $\mathbf{R}_{\scriptscriptstyle N11}$ represents its information factor and $\mathbf{R}_{\scriptscriptstyle N12}$ the cross information between the two maps.
\subsection{Exploration: General Case}\label{ssec:newexp}
In general, when in exploration mode, past map's states (from previous exploration and relocalization phases) do not need to be updated. Thus, we apply RISE with $\mathbf{x}_\two = \mathbf{x}_{\scriptscriptstyle M}^{\prime}$ being the past map's states, while $\mathbf{x}_\one = \mathbf{x}_{\scriptscriptstyle N}$ comprises the new map's states which expand as the current exploration goes on.
%
% When in exploration mode, the system not only estimates the states of the new map built in the current exploration (see Sec.~\ref{ssec:exploration1}), but also maintains their correlations with the old map's states.
%
Specifically, the same operations as for the first exploration (Sec.~\ref{ssec:exploration1}) are applied to the \emph{new} map's states.
Additionally, as explained in Sec.~\ref{ssec:schmidt}, the RISE applies the QR transformation in~\eqref{eq:qr_first_exploration} on the cross information term $\mathbf{F}_{\scriptscriptstyle E}$ (originally starts as $\mathbf{R}_{\scriptscriptstyle N12}$ in~\eqref{eq:cn} and evolves with time), as shown in Fig.~\ref{fig:exp_new1}, and drops the cost term $\left\|\mathbf{F}_{\scriptscriptstyle E3}^\oplus\left(\mathbf{x}_{\scriptscriptstyle M}^\prime-\hat{\mathbf{x}}_{\scriptscriptstyle M}^\prime\right)-\mathbf{e}_{\scriptscriptstyle E}\right\|^2$ involving only the \emph{old} map's states (see~\cite{risetechreport} for a detailed derivation). Note that only the lower of part of $\mathbf{F}_{\scriptscriptstyle E}$, $\mathbf{F}_{\scriptscriptstyle E2}$, which has a bounded number of dense columns, needs updating.
Thus, the computational complexity for the general case of exploration with RISE is also \emph{constant}.

\section{RISE-SLAM: Relocalization}\label{sec:relocalization}
We now consider the case, where the system switches from exploration to relocalization mode so as to use loop-closure measurements for global pose and map correction.
\subsection{Transition from Exploration to Relocalization}\label{ssec:torelo}
Before entering relocalization, we need to switch the state order from \emph{chronological} to \emph{reverse chronological}.
Specifically, during exploration, the state vector is in the form $\begin{bmatrix}\mathbf{x}_{\scriptscriptstyle L}^{\T} & \mathbf{x}_{\scriptscriptstyle M}^{\prime\T}\end{bmatrix}^\T$,
where $\mathbf{x}_{\scriptscriptstyle L}$ comprises all states in the current map, while $\mathbf{x}_{\scriptscriptstyle M}^{\prime}$ corresponds to the old map [see~\eqref{eq:cn} and Sec.~\ref{ssec:newexp}].
We first change the state order of $\mathbf{x}_{\scriptscriptstyle L}$ to reverse chronological by defining $\mathbf{x}_{\scriptscriptstyle L}^\prime\triangleq\mathbf{P}_{\scriptscriptstyle L}\mathbf{x}_{\scriptscriptstyle L}$ ($\mathbf{P}_{\scriptscriptstyle L}$ is a permutation matrix).
Then, we split $\mathbf{x}_{\scriptscriptstyle L}^\prime$ into two parts: $\mathbf{x}_{\scriptscriptstyle L1}^\prime$ comprises the \emph{recent} states to be maintained in the frontend, while the remaining states $\mathbf{x}_{\scriptscriptstyle L2}^\prime$ are combined with the old map $\mathbf{x}_{\scriptscriptstyle M}^\prime$ so as to be optimized by the backend ($\mathbf{x}_{\scriptscriptstyle B}^\prime\triangleq\begin{bmatrix}\mathbf{x}_{\scriptscriptstyle L2}^{\prime\T} & \mathbf{x}_{\scriptscriptstyle M}^{\prime\T}\end{bmatrix}^\T$).
Accordingly, in terms of cost functions, we first combine all available cost terms so far, including the current exploration (Fig.~\ref{fig:exp_new1}) and the old map $\mathcal{C}_{\scriptscriptstyle M}$ from~\eqref{eq:cn} [first term in~\eqref{eq:ct}], and the new loop-closure measurements [second term in~\eqref{eq:ct}], to get the total cost $\mathcal{C}_{\scriptscriptstyle T}$.
Then, we employ the QR factorization of RISE [see~\eqref{eq:qr1}-\eqref{eq:qr1_rhs}] to split $\mathcal{C}_{\scriptscriptstyle T}$ into two parts:
\begin{align}
    \hspace*{-0.5mm}\mathcal{C}_{\scriptscriptstyle T}&=\left\|\begin{bmatrix}
    \mathbf{R}_{\scriptscriptstyle T11} & \mathbf{R}_{\scriptscriptstyle T12} \\
    & \mathbf{R}_{\scriptscriptstyle T22}
    \end{bmatrix}\begin{bmatrix}\mathbf{x}_{\scriptscriptstyle L1}^\prime-\hat{\mathbf{x}}_{\scriptscriptstyle L1}^\prime\\\mathbf{x}_{\scriptscriptstyle B}^\prime-\hat{\mathbf{x}}_{\scriptscriptstyle B}^\prime\end{bmatrix}
    \right\|^2\nonumber\\&\ \ \ +\left\|
    \mathbf{H}_{\scriptscriptstyle T1}(\mathbf{x}_{\scriptscriptstyle L1}^\prime-\hat{\mathbf{x}}_{\scriptscriptstyle L1}^\prime)+\mathbf{H}_{\scriptscriptstyle T2}(\mathbf{x}_{\scriptscriptstyle B}^\prime-\hat{\mathbf{x}}_{\scriptscriptstyle B}^\prime)
    -\mathbf{r}_{\scriptscriptstyle T}
    \right\|^2\label{eq:ct}\\
    % \end{align}\begin{align}
    % \mathcal{C}_{\scriptscriptstyle T}&=\left\|\begin{bmatrix}
    &=\left\|\begin{bmatrix}
    \mathbf{R}_{\scriptscriptstyle T11}^\oplus & \mathbf{R}_{\scriptscriptstyle T12}^\oplus \\
    & \mathbf{R}_{\scriptscriptstyle T22}\\
    & \mathbf{H}_{\scriptscriptstyle T2}^\oplus
    \end{bmatrix}\begin{bmatrix}\mathbf{x}_{\scriptscriptstyle L1}^\prime-\hat{\mathbf{x}}_{\scriptscriptstyle L1}^\prime\\\mathbf{x}_{\scriptscriptstyle B}^\prime-\hat{\mathbf{x}}_{\scriptscriptstyle B}^\prime\end{bmatrix}
    -\begin{bmatrix}\mathbf{r}_{\scriptscriptstyle T}^\oplus\\\mathbf{0}\\\mathbf{e}_{\scriptscriptstyle T}\end{bmatrix}\right\|^2
    =\mathcal{C}_{\scriptscriptstyle F}+\mathcal{C}_{\scriptscriptstyle B}
\label{eq:cfb}\end{align}
where
    \begin{align}
    \mathcal{C}_{\scriptscriptstyle F}&\triangleq
    % \left\|\begin{bmatrix}
    % \mathbf{R}_{\scriptscriptstyle T11}^\oplus & \mathbf{R}_{\scriptscriptstyle T12}^\oplus
    % \end{bmatrix}\begin{bmatrix}
    % \mathbf{x}_{\scriptscriptstyle L1}^\prime-\hat{\mathbf{x}}_{\scriptscriptstyle L1}^{\prime\oplus}\\
    % \mathbf{x}_{\scriptscriptstyle B}^\prime-\hat{\mathbf{x}}_{\scriptscriptstyle B}^\prime\\
    % \end{bmatrix}\right\|^2\\
    \left\|\mathbf{R}_{\scriptscriptstyle T11}^\oplus(\mathbf{x}_{\scriptscriptstyle L1}^\prime-\hat{\mathbf{x}}_{\scriptscriptstyle L1}^{\prime\oplus})+ \mathbf{R}_{\scriptscriptstyle T12}^\oplus(\mathbf{x}_{\scriptscriptstyle B}^\prime-\hat{\mathbf{x}}_{\scriptscriptstyle B}^\prime)\right\|^2\label{eq:cf}\\
    \hat{\mathbf{x}}_{\scriptscriptstyle L1}^{\prime\oplus}&\triangleq\hat{\mathbf{x}}_{\scriptscriptstyle L1}^{\prime}+\mathbf{R}_{\scriptscriptstyle T11}^{\oplus-1}\mathbf{r}_{\scriptscriptstyle T}^\oplus\\
    \mathcal{C}_{\scriptscriptstyle B}&\triangleq
    \left\|\begin{bmatrix}
    \mathbf{R}_{\scriptscriptstyle T22}\\
    \mathbf{H}_{\scriptscriptstyle T2}^\oplus
    \end{bmatrix}
    (\mathbf{x}_{\scriptscriptstyle B}^\prime-\hat{\mathbf{x}}_{\scriptscriptstyle B}^\prime)-\begin{bmatrix}
        \mathbf{0}\\
    \mathbf{e}_{\scriptscriptstyle T}
    \end{bmatrix}\right\|^2\label{eq:cb}
\end{align}

Among the operations required in this transition step, the frontend only needs a small-size matrix QR factorization to obtain $\mathcal{C}_{\scriptscriptstyle F}$ with \emph{constant} cost, while all the (expensive) remaining work is done in the backend (see~\cite{risetechreport} for details).

After this transition step, in the \emph{frontend} $\mathcal{C}_{\scriptscriptstyle F}$ is extended to include information from new poses and features and minimized to update a sliding window of recent states (starting from $\mathbf{x}_{\scriptscriptstyle L1}^\prime$).
Meanwhile, in the \emph{backend}, $\mathcal{C}_{\scriptscriptstyle B}$ is minimized to provide global corrections for the past states $\mathbf{x}_{\scriptscriptstyle B}^\prime$.

\subsection{Relocalization: Frontend Thread}\label{ssec:frontend}
\begin{figure}
    \center
    \includegraphics[width=0.45\textwidth]{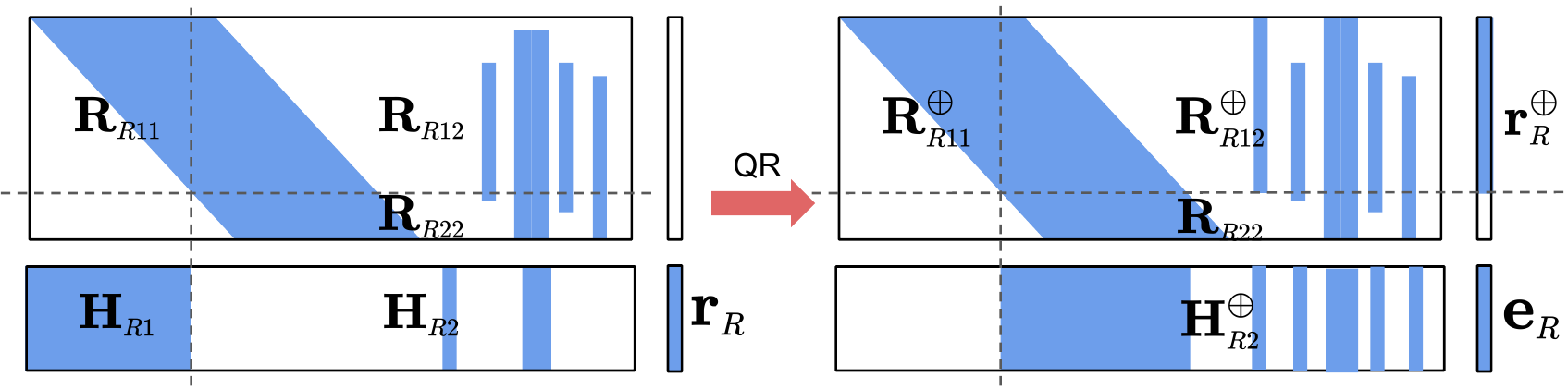}
    \vspace{-1mm}
    \caption{Structure of the factors corresponding to the relocalization cost terms before and after a RISE update in the frontend.}
    \label{fig:rv1}
    \vspace{-5mm}
\end{figure}

The relocalization frontend employs RISE to process both local feature tracks and loop-closure measurements so as to update a sliding window of \emph{recent} states (see \textcircled{\scriptsize 3} in Fig.~\ref{fig:overview}). Specifically, the cost function $\mathcal{C}_{\scriptscriptstyle R}$, to be optimized in the frontend, extends from $\mathcal{C}_{\scriptscriptstyle F}$ in~\eqref{eq:cf} and has the form
\begin{align}
    \mathcal{C}_{\scriptscriptstyle R}&=\left\|\begin{bmatrix}
    \mathbf{R}_{\scriptscriptstyle R11} & \mathbf{R}_{\scriptscriptstyle R12}\\
     & \mathbf{R}_{\scriptscriptstyle R22}
    \end{bmatrix}
    \begin{bmatrix}
    \mathbf{x}_{\scriptscriptstyle R1}^\prime-\hat{\mathbf{x}}_{\scriptscriptstyle R1}^\prime\\
    \mathbf{x}_{\scriptscriptstyle R2}^\prime-\hat{\mathbf{x}}_{\scriptscriptstyle R2}^\prime
    \end{bmatrix}
    \right\|^2\nonumber\\&\ \ \ +
    \left\|\begin{bmatrix}\mathbf{H}_{\scriptscriptstyle R1} & \mathbf{H}_{\scriptscriptstyle R2}\end{bmatrix}
    \begin{bmatrix}\mathbf{x}_{\scriptscriptstyle R1}^\prime-\hat{\mathbf{x}}_{\scriptscriptstyle R1}^\prime\\
    \mathbf{x}_{\scriptscriptstyle R2}^\prime-\hat{\mathbf{x}}_{\scriptscriptstyle R2}^\prime\end{bmatrix}-\mathbf{r}_{\scriptscriptstyle R}
    \right\|^2\label{eq:cr_before_qr}
\end{align}
where $\mathbf{x}_{\scriptscriptstyle R1}^\prime$ contains the window of recent states to be updated at the current step, while $\mathbf{x}_{\scriptscriptstyle R2}^\prime$ represents all previous states (including $\mathbf{x}_{\scriptscriptstyle B}^\prime$) whose estimates are kept unchanged in the frontend.
%contains states being optimized in the backend ($\mathbf{x}_{\scriptscriptstyle B}^\prime$) as well as past states added to the state vector by the frontend after the backend started.
The measurement Jacobians $\mathbf{H}_{\scriptscriptstyle R1}$ and $\mathbf{H}_{\scriptscriptstyle R2}$ are both nonzero (see Fig.~\ref{fig:rv1}) since here we consider both local feature tracks and loop-closure measurements.
%$\mathcal{C}_{\scriptscriptstyle B}$ is the cost function minimized by the backend and not used in the frontend.
Then, the frontend employs RISE to update only $\mathbf{x}_{\scriptscriptstyle R1}^\prime$ by first performing the QR factorization to transform $\mathcal{C}_{\scriptscriptstyle R}$ from~\eqref{eq:cr_before_qr} into
\begin{align}
    \mathcal{C}_{\scriptscriptstyle R}%&=\left\|\begin{bmatrix}
    %\mathbf{R}_{\scriptscriptstyle R11} & \mathbf{R}_{\scriptscriptstyle R12}\\
    %\mathbf{H}_{\scriptscriptstyle R1} & \mathbf{H}_{\scriptscriptstyle R2}
    %\end{bmatrix}
    %\begin{bmatrix}
    %\mathbf{x}_{\scriptscriptstyle R1}^\prime-\hat{\mathbf{x}}_{\scriptscriptstyle R1}^\prime\\
    %\mathbf{x}_{\scriptscriptstyle R2}^\prime-\hat{\mathbf{x}}_{\scriptscriptstyle R2}^\prime
    %\end{bmatrix}-\begin{bmatrix}\mathbf{0}\\\mathbf{r}_{\scriptscriptstyle R}\end{bmatrix}
    %\right\|^2\nonumber\\&+
    %\left\|\mathbf{R}_{\scriptscriptstyle R22}(\mathbf{x}_{\scriptscriptstyle R2}^\prime-\hat{\mathbf{x}}_{\scriptscriptstyle R2}^\prime)
    %\right\|^2+\mathcal{C}_{\scriptscriptstyle B}\\
    &=\left\|\begin{bmatrix}
    \mathbf{R}_{\scriptscriptstyle R11}^\oplus & \mathbf{R}_{\scriptscriptstyle R12}^\oplus\\
    & \mathbf{H}_{\scriptscriptstyle R2}^\oplus
    \end{bmatrix}
    \begin{bmatrix}
    \mathbf{x}_{\scriptscriptstyle R1}^\prime-\hat{\mathbf{x}}_{\scriptscriptstyle R1}^\prime\\
    \mathbf{x}_{\scriptscriptstyle R2}^\prime-\hat{\mathbf{x}}_{\scriptscriptstyle R2}^\prime
    \end{bmatrix}-\begin{bmatrix}\mathbf{r}_{\scriptscriptstyle R}^\oplus\\\mathbf{e}_{\scriptscriptstyle R}\end{bmatrix}
    \right\|^2\nonumber\\&\ \ \ +
    \left\|\mathbf{R}_{\scriptscriptstyle R22}(\mathbf{x}_{\scriptscriptstyle R2}^\prime-\hat{\mathbf{x}}_{\scriptscriptstyle R2}^\prime)\right\|^2\label{eq:cr_after_qr}
    \end{align}
and then dropping the term $\left\|\mathbf{H}_{\scriptscriptstyle R2}^\oplus(\mathbf{x}_{\scriptscriptstyle R2}^\prime-\hat{\mathbf{x}}_{\scriptscriptstyle R2}^\prime)-\mathbf{e}_{\scriptscriptstyle R}\right\|^2$ (since we are not updating $\mathbf{x}_{\scriptscriptstyle R2}^\prime$) to get $\bar{\mathcal{C}}_{\scriptscriptstyle R}$
\begin{align}
    \bar{\mathcal{C}}_{\scriptscriptstyle R}&=\left\|\begin{bmatrix}
    \mathbf{R}_{\scriptscriptstyle R11}^\oplus & \mathbf{R}_{\scriptscriptstyle R12}^\oplus\\
    & \mathbf{R}_{\scriptscriptstyle R22}
    \end{bmatrix}
    \begin{bmatrix}
    \mathbf{x}_{\scriptscriptstyle R1}^\prime-\hat{\mathbf{x}}_{\scriptscriptstyle R1}^{\prime\oplus}\\
    \mathbf{x}_{\scriptscriptstyle R2}^\prime-\hat{\mathbf{x}}_{\scriptscriptstyle R2}^\prime
    \end{bmatrix}\right\|^2\label{eq:crbar}\\
    \hat{\mathbf{x}}_{\scriptscriptstyle R1}^{\prime\oplus}&\triangleq\hat{\mathbf{x}}_{\scriptscriptstyle R1}^{\prime}+\mathbf{R}_{\scriptscriptstyle R11}^{\oplus-1}\mathbf{r}_{\scriptscriptstyle R}^\oplus
\end{align}
where $\hat{\mathbf{x}}_{\scriptscriptstyle R1}^{\prime\oplus}$ is the updated state estimate.
The structural changes of the factors appearing in~\eqref{eq:cr_before_qr}-\eqref{eq:cr_after_qr} are shown in Fig.~\ref{fig:rv1}.
The processing cost of this update is defined by the QR factorization, and is \emph{constant} since the size of $\mathbf{x}_{\scriptscriptstyle R1}^\prime$ and the number of dense columns in $\mathbf{R}_{\scriptscriptstyle R12}^\oplus$ are bounded.
Due to its low processing cost, the frontend thread is able to update the states at high frequency in real time.
%%%%%%%%%%%%%%%%%%%%%%%%%%%%%%%%%%%%%%%%%%%%%%%%%%%%%%%%%%%%%%%%%%%%%%%%%%%%%%
\subsection{Relocalization: Backend Thread}\label{ssec:backend}
The backend performs global adjustment so as to accurately update a large number of \emph{past} states, while running in parallel with the frontend to avoid blocking it (see \textcircled{\scriptsize 5} in Fig.~\ref{fig:overview}).
Within RISE, the backend can run whenever the thread is idle, to increase accuracy. In our current design, however, in order to save processing, we choose to run it only once during each relocalization phase, right after the transition step described in Sec.~\ref{ssec:torelo}, and update all the past states to obtain the optimal solution.
% $\left\|\mathbf{H}_{\scriptscriptstyle T}^\oplus(\mathbf{x}_{\scriptscriptstyle B}^\prime-\hat{\mathbf{x}}_{\scriptscriptstyle B}^\prime)-\mathbf{e}_{\scriptscriptstyle T}\right\|^2$
Specifically, from~\eqref{eq:cb}, we compute the optimal estimate of the past states $\mathbf{x}_{\scriptscriptstyle B}^\prime$ by minimizing the cost function $\mathcal{C}_{\scriptscriptstyle B}$, which is a batch least-squares problem and can be solved by employing a sparse QR factorization.
Once the new estimate $\hat{\mathbf{x}}_{\scriptscriptstyle B}^{\prime\oplus}$ is obtained, $\mathcal{C}_{\scriptscriptstyle B}$ can be rewritten as (after ignoring a constant term)
\begin{align}
    \mathcal{C}_{\scriptscriptstyle B}=\left\|\mathbf{R}_{\scriptscriptstyle B}(\mathbf{x}_{\scriptscriptstyle B}^\prime-\hat{\mathbf{x}}_{\scriptscriptstyle B}^{\prime\oplus})
    \right\|^2\label{eq:cb2}
\end{align}
The computational cost is approximately \emph{linear} in the (large) size of $\mathbf{x}_{\scriptscriptstyle B}^\prime$, but it does not prevent the frontend from performing real-time estimation since they run in parallel.
%%%%%%%%%%%%%%%%%%%%%%%%%%%%%%%%%%%%%%%%%%%%%%%%%%%%%%%%%%%%%%%%%%%%%%%%%%%%%%
\subsection{Feedback from Backend to Frontend Thread}\label{ssec:back2front}
After the backend finishes a global update on the \emph{past} states $\mathbf{x}_{\scriptscriptstyle B}^\prime$, the frontend employs this result to update the \emph{recent} states so that they benefit from the global correction (see \textcircled{\scriptsize 4} in Fig.~\ref{fig:overview}).
Specifically, if we denote all the recent states accumulated in the frontend since the backend started as $\mathbf{x}_{\scriptscriptstyle F}^\prime$, then from~\eqref{eq:crbar} and Fig.~\ref{fig:rv1}, the frontend maintains a cost term of the form $\left\|\mathbf{R}_{\scriptscriptstyle F}(\mathbf{x}_{\scriptscriptstyle F}^\prime-\hat{\mathbf{x}}_{\scriptscriptstyle F}^\prime)+ \mathbf{R}_{\scriptscriptstyle FB}(
\mathbf{x}_{\scriptscriptstyle B}^\prime-\hat{\mathbf{x}}_{\scriptscriptstyle B}^\prime)
\right\|^2$, where the information factor $\mathbf{R}_F$ is upper-triangular and $\mathbf{R}_{FB}$ represents the cross information between the states in the frontend and the backend.
Combining this frontend's cost term with that of the backend [$\mathcal{C}_{\scriptscriptstyle B}$ in~\eqref{eq:cb2}] yields
\begin{align}
    \mathcal{C}_{\scriptscriptstyle FB}&=\left\|\mathbf{R}_{\scriptscriptstyle F}(\mathbf{x}_{\scriptscriptstyle F}^\prime-\hat{\mathbf{x}}_{\scriptscriptstyle F}^\prime)+ \mathbf{R}_{\scriptscriptstyle FB}(
    \mathbf{x}_{\scriptscriptstyle B}^\prime-\hat{\mathbf{x}}_{\scriptscriptstyle B}^\prime)
    \right\|^2\nonumber\\
    &\ \ \ +\left\|\mathbf{R}_{\scriptscriptstyle B}(\mathbf{x}_{\scriptscriptstyle B}^\prime-\hat{\mathbf{x}}_{\scriptscriptstyle B}^{\prime\oplus})
    \right\|^2\nonumber\\
    &=\left\|\begin{bmatrix}
    \mathbf{R}_{\scriptscriptstyle F} & \mathbf{R}_{\scriptscriptstyle FB}\\
    & \mathbf{R}_{\scriptscriptstyle B}
    \end{bmatrix}
    \begin{bmatrix}\mathbf{x}_{\scriptscriptstyle F}^\prime-\hat{\mathbf{x}}_{\scriptscriptstyle F}^{\prime\oplus}\\
    \mathbf{x}_{\scriptscriptstyle B}^\prime-\hat{\mathbf{x}}_{\scriptscriptstyle B}^{\prime\oplus}\end{bmatrix}
    \right\|^2\label{eq:cfb_feedback}\\
    \hat{\mathbf{x}}_{\scriptscriptstyle F}^{\prime\oplus}&\triangleq\hat{\mathbf{x}}_{\scriptscriptstyle F}^\prime+\mathbf{R}_{\scriptscriptstyle F}^{-1}\mathbf{R}_{\scriptscriptstyle FB}
    \left(\hat{\mathbf{x}}_{\scriptscriptstyle B}^\prime-\hat{\mathbf{x}}_{\scriptscriptstyle B}^{\prime\oplus}\right)\label{eq:xf_update_feedback}
\end{align}
where $\hat{\mathbf{x}}_{\scriptscriptstyle F}^{\prime\oplus}$ is the globally corrected estimate for the recent states in the frontend. The operations required in~\eqref{eq:xf_update_feedback} are sparse matrix-vector multiplications and back substitutions, which are faster than the operations of Sec.~\ref{ssec:frontend} in practice.

As a result of this step, all current states in the frontend are \emph{immediately} corrected using the globally adjusted estimates from the backend.
Note that, this is a key difference between our algorithm and existing multi-thread SLAM systems. Specifically, \cite{mur2015orb,qin2018vins,liu2018ice} solve \emph{separate} optimization problems independently in different threads; hence their frontend has to rely on map reobservations \emph{after} the backend's global adjustment finishes in order to obtain corrections. This is due to the fact that the two sets of states involved in their frontend's and backend's (\emph{separate}) optimization problems are considered \emph{uncorrelated}.
In contrast, our algorithm always solves a \emph{single} optimization problem over two \emph{correlated} sets of states [see~\eqref{eq:cfb_feedback}], carried out in two separate threads. Consequently, as soon as the corrections from the backend become available, they immediately affect the frontend's states, even if the map is not reobserved.

%And this feedback process can always be carried out, even if the camera has already entered a new area and started a new exploration phase, due to the fact that we are solving a \emph{single} optimization problem (in two separate threads) as mentioned before. This is in contrast to all existing multi-thread SLAM systems, such as~\cite{mur2015orb,qin2018vins,liu2018ice}, that solve \emph{multiple} optimization problems independently in different threads. In those systems, the frontend has to rely on subsequent map reobservations to obtain information from the backend.

\section{Experimental Results}
\label{sec:experimental_results}
To evaluate the performance of the proposed RISE-SLAM, we compared it against state-of-the-art visual-inertial SLAM estimators, including a visual-inertial odometry (VIO) system without loop closures (OKVIS~\cite{KFSLAMJ}), and SLAM systems with loop closures (VINS-Mono~\cite{qin2018vins}, ICE-BA~\cite{liu2018ice}), using the EuRoC~\cite{Burri25012016} datasets. Since our implementation focuses on visual-inertial SLAM, we did not compare to vision-only systems, such as~\cite{mur2015orb}. The datasets contain stereo images (only the left-camera images are used) from global shutter cameras (20 Hz) and IMU measurements (200 Hz), along with ground-truth poses from VICON. The code of each compared system is downloaded from their Github repositories and run with the provided configuration files for the EuRoC datasets.
\subsection{System Setup}
In our implementation, we extract 300 ORB~\cite{rublee2011orb} features per image and match them based on their descriptors  against previous images to generate feature tracks. Loop-closure measurements are provided by a vocabulary tree~\cite{nister2006scalable}, and the system switches to relocalization mode when detecting features that have not been observed for more than 15 sec. We model the noise of all visual observations as zero-mean white Gaussian with $\sigma=1.5$ pixels.
During both exploration and relocalization, the frontend updates a sliding window corresponding to 10 recent poses and their observed features.
To improve efficiency, feature tracks longer than 20 are split into multiple shorter ones while the number of feature tracks being processed per time step is limited to 40.
%
%Due to the superior numerical stability of square root estimators, we employ single-precision floating numbers to achieve fast computation.
%
\subsection{Localization Accuracy}\label{ssec:accuracy}
We compute the root-mean-square error (RMSE) of the estimated positions as compared to the ground truth, to evaluate the tracking accuracy of all estimators considered. Each estimated trajectory is aligned with the ground-truth coordinate frame by a 3D-to-3D matching using~\cite{horn1987closed}.

For fairness, we first run the proposed RISE-SLAM without using any loop-closure measurements [RISE (vio)] so as to compare it with the VIO system OKVIS. Then, we add loop-closure measurements [RISE (lc)] to compare it with the SLAM systems VINS-Mono and ICE-BA. The position RMSE results from all EuRoC datasets are shown in Table~\ref{tab:rmse}.
As shown, our RISE (vio) outperforms OKVIS on all datasets.
This is due to the fact that RISE optimally processes all available measurements during exploration (see Sec.~\ref{ssec:exploration1}). Among the SLAM systems, ICE-BA performs the worst. This is probably because loop closure is not implemented yet in their public code, which makes their system a VIO with a backend that performs global smoothing. Regardless, ICE-BA does not achieve better accuracy even when compared to RISE (vio), while requiring more CPU resources for running a backend thread. Meanwhile, our RISE (lc) is the best on most of the datasets, and outperforms VINS-Mono. An example of the RISE-SLAM estimated trajectory vs. ground truth is shown in the accompanying video.%\footnote{RISE-SLAM performs consistently well on VICON room (V) datasets, while a little worse on machine hall (MH) datasets. We believe this is due to our policy for accepting loop-closure measurements. We are currently investigating potential improvements on robustness to outliers.} %Fig.~\ref{fig:m5} shows two example of estimated trajectories against ground truth.

\begin{table}
    \centering
    \caption{Position RMSE (cm) on EuRoC datasets}
    \vspace{-2mm}
    \setlength{\tabcolsep}{5pt}
    {\scriptsize
    \begin{tabular}{c|cc|ccc}
    \hline
    Dataset  & RISE (vio) & OKVIS & RISE (lc)  & VINS-Mono & ICE-BA\\\hline
%             & w/o lc  &  & w/ lc & &\\\hline
MH\_01\_easy & \textbf{16.1} & 34.6 & \textbf{5.1} & 8.8 & 16.4\\
MH\_02\_easy & \textbf{19.3} & 40.9 & 12.0 & \textbf{6.3} & 8.9\\
MH\_03\_medium & \textbf{21.9} & 22.1 & \textbf{8.4} & 8.5 & 17.2\\
MH\_04\_difficult & \textbf{24.1} & 33.7 & \textbf{16.1} & 17.5 & 31.6\\
MH\_05\_difficult & \textbf{30.5} & 42.5 & 25.6 & \textbf{14.5} & 31.4\\
V1\_01\_easy & \textbf{5.1} & 10.6 & \textbf{4.1} & 4.5 & 5.2\\
V1\_02\_medium & \textbf{7.0} & 11.6 & \textbf{3.3} & 6.5 & 11.7\\
V1\_03\_difficult & \textbf{11.6} & 22.7 & \textbf{6.7} & 29.9 & 11.1\\
V2\_01\_easy & \textbf{6.6} & 16.2 & \textbf{5.5} & 6.4 & 9.1\\
V2\_02\_medium & \textbf{10.4} & 18.1 & \textbf{4.0} & 13.3 & 11.5\\
V2\_03\_difficult & \textbf{10.1} & 28.5 & \textbf{11.7} & 18.0 & 12.8\\\hline
Average & \textbf{14.8} & 25.6 & \textbf{9.3} & 12.2 & 15.2\\\hline
% MH\_01\_easy & \textbf{11.5}  & 34.6 & \textbf{6.3} & 8.8 & 16.4\\
% MH\_02\_easy & \textbf{19.2}  & 40.9 & 17.9 & \textbf{6.3} & 8.9\\
% MH\_03\_medium & 28.7 & \textbf{22.1} & 20.3 & \textbf{8.5} & 17.2\\
% MH\_04\_difficult & \textbf{20.0} & 33.7 & \textbf{15.7} & 17.5 & 31.6\\
% MH\_05\_difficult & \textbf{38.2} & 42.5 & \textbf{14.1} & 14.5 & 31.4\\
% V1\_01\_easy & \textbf{7.7} & 10.6 & \textbf{4.1} & 4.5 & 5.2\\
% V1\_02\_medium & \textbf{9.6} & 11.6 & \textbf{6.0} & 6.5 & 11.7\\
% V1\_03\_difficult & \textbf{10.4} & 22.7 & \textbf{9.3} & 29.9 & 11.1\\
% V2\_01\_easy & \textbf{6.8} & 16.2 & \textbf{5.9} & 6.4 & 9.1\\
% V2\_02\_medium & \textbf{12.3} & 18.1 & \textbf{5.2} & 13.3 & 11.5\\
% V2\_03\_difficult & \textbf{12.0} & 28.5 & \textbf{9.7} & 18.0 & 12.8\\\hline
% Average & \textbf{16.0} & 25.6 & \textbf{10.4} & 12.2 & 15.2\\\hline 
    % MH\_01\_easy & \textbf{0.063} &0.088 &0.164 & &0.346\\
    % MH\_02\_easy & 0.179 & \textbf{0.063} &0.089 & &0.409\\
    % MH\_03\_medium & 0.203 & \textbf{0.085} &0.172& &0.221\\
    % MH\_04\_difficult & \textbf{0.157} &0.175 &0.316& &0.337 \\
    % MH\_05\_difficult & \textbf{0.141} &0.145 &0.314& &0.425 \\
    % V1\_01\_easy & \textbf{0.041} &0.045 &0.052& &0.106 \\
    % V1\_02\_medium & \textbf{0.060} &0.065 &0.117& &0.116 \\
    % V1\_03\_difficult & \textbf{0.093} &0.299 &0.111& &0.227 \\
    % V2\_01\_easy & \textbf{0.059} &0.064 &0.091& &0.162 \\
    % V2\_02\_medium & \textbf{0.052} &0.133 &0.115& &0.181 \\
    % V2\_03\_difficult & \textbf{0.097} &0.180 &0.128& &0.285 \\\hline
    % average & \textbf{0.104} &0.122 &0.152& &0.256 \\\hline
    \end{tabular}
    }
    \label{tab:rmse}
    \vspace{-1mm}
\end{table}

% \begin{figure}
%     \center
%     \includegraphics[width=0.25\textwidth,valign=t]{fig/v22.pdf}\includegraphics[width=0.24\textwidth,valign=t]{fig/m5.pdf}
%     \caption{Comparison of estimated trajectories versus ground truth on EuRoC datasets V2\_02\_medium and MH\_05\_difficult}
%     \label{fig:m5}
% \end{figure}

\subsection{Computational Efficiency}
We run all algorithms on a laptop with Intel Core i7-6700HQ 2.60GHz x 8 CPU to compare the efficiency of each estimator. Since we focus on estimation speed, we do not consider the cost of image processing. Average times of one run are shown in Table~\ref{tab:time}. Note that since RISE-SLAM has two modes with different costs, we report the averages of exploration and relocalization (the time of transition steps is also included), respectively, as well as the overall average across all runs. As expected, the exploration is typically faster than relocalization since it only processes local feature tracks. As evident from Table~\ref{tab:time}, RISE-SLAM is significantly faster than VINS-Mono and OKVIS. Only ICE-BA runs as fast as RISE-SLAM, but has lower tracking accuracy.\footnote{Although we did not compare to ORB-SLAM as they do not use IMU, it is worth noting that as reported in~\cite{liu2018ice} ORB-SLAM is more than 10 times slower than~\cite{liu2018ice} whose average processing time is almost the same as ours.}
%
%\textcolor{blue}{ORB-SLAM is much slower than ours which has similar speed to~\cite{liu2018ice}, while the ORB-SLAM is more than 10$\times$ slower than~\cite{liu2018ice} as reported in~\cite{liu2018ice}.}
%
%Overall, the proposed estimator outperforms other alternatives by achieving both high tracking accuracy and low computational cost at the same time.
%
\begin{table}
    \centering
    \caption{Average running time (msec) of one estimator run}
    \vspace{-2mm}
    {\scriptsize
    \begin{tabular}{cccccc}
    \hline
    RISE & RISE & RISE & OKVIS & VINS-Mono & ICE-BA \\
    exploration & relocalization & overall & & & \\\hline
    10 & 13 & 11 & 26 & 52 & 11 \\\hline
    \end{tabular}
    }
    \label{tab:time}
    \vspace{-5mm}
\end{table}

\subsection{Estimation Consistency}
To assess the consistency improvement of RISE as compared to estimators that assume some past states to be perfectly known, we employ the normalized estimation error squared (NEES)~\cite{bar2004estimation}. Typically more overconfident (inconsistent) estimates yield larger NEES. %We first do comparison using a visual-inertial simulation, where there are less other factors affecting consistency (e.g., unmodeled errors and outliers). The simulated trajectory is a disturbed circle with two loops, where the camera observes new features during the first loop and reobserve them during the second. Fig.~\ref{fig:nees_sim} shows that comparing to ignoring the uncertainty of past states, applying RISE has much smaller position NEES and hence improves consistency.
%
% Additionally, we compare the position NEES of RISE-SLAM with VINS-Mono (whose covariance of VINS-Mono's estimate is got from Ceres, the optimization library it uses), which assuming key frame poses as perfectly known in relocalization update. \textcolor{blue}{On EuRoC datasets, VINS-Mono always shows a median position NEES of 100+ while ours typically yield 30-50 (see~\cite{risetechreport} for detailed results).} Hence, RISE-SLAM yields better consistency than VINS-Mono, which employs inconsistent approximation, even on real datasets.
%In addition, we manage to run our pipeline on PIXEL 1 cell phone, and get similar timings to Table.~\ref{tab:time}.
To isolate the effect of other factors affecting consistency (e.g., linearization, local minima, outliers), we first evaluate NEES in Monte Carlo simulations where the camera follows a circular path twice. In this case, the RISE-SLAM's position NEES fluctuates between 5 and 100, while if the map is assumed perfectly-known the NEES ranges from 10 to 2,500. Furthermore, on the EuRoC datasets, VINS-Mono, which assumes perfectly-known keyframe poses, has a median position NEES of over 100, while that of RISE-SLAM is $\sim$40 (see~\cite{risetechreport} for details).

% \begin{figure}
%     \centering
%     \includegraphics[width=0.3\textwidth]{fig/nees_sim.eps}
%     \caption{NEES comparison in simulation: We run RISE-SLAM exploration mode during the first loop, and for the second loop where there are loop-closure measurements, we employ (a) RISE-SLAM relocalization mode (b) assuming past poses and features as perfectly known (i.e. zeroing out Jacobian w.r.t them).}
%     \label{fig:nees_sim}
% \end{figure}
\section{Conclusion}
Motivated by the consistency guarantees and the linear processing cost of the Schmidt Kalman filter, as well as the linear memory requirements of the Cholesky factor of maps computed online, in this work, we introduce the resource-aware inverse Schmidt estimator (RISE), an approximation to the exact inverse Schmidt, that maintains the sparsity of the Cholesky factor and allows us to improve consistency.
Furthermore, we employ different (in terms of the order and number of states involved) configurations of RISE to form the frontend and backend of a visual-inertial system, RISE-SLAM, which appropriately treats the exploration versus the relocalization phases of SLAM to achieve real-time operation.
As demonstrated by the experimental evaluation, RISE-SLAM achieves position accuracy typically better than alternative state-of-the-art visual-inertial estimators, at lower processing cost while improving estimation consistency.

%\addtolength{\textheight}{-12cm}   % This command serves to balance the column lengths
                                  % on the last page of the document manually. It shortens
                                  % the textheight of the last page by a suitable amount.
                                  % This command does not take effect until the next page
                                  % so it should come on the page before the last. Make
                                  % sure that you do not shorten the textheight too much.

%%%%%%%%%%%%%%%%%%%%%%%%%%%%%%%%%%%%%%%%%%%%%%%%%%%%%%%%%%%%%%%%%%%%%%%%%%%%%%%%

\bibliographystyle{IEEEtran}
\bibliography{references}

\begin{thebibliography}{10}
\providecommand{\url}[1]{#1}
\csname url@rmstyle\endcsname
\providecommand{\newblock}{\relax}
\providecommand{\bibinfo}[2]{#2}
\providecommand\BIBentrySTDinterwordspacing{\spaceskip=0pt\relax}
\providecommand\BIBentryALTinterwordstretchfactor{4}
\providecommand\BIBentryALTinterwordspacing{\spaceskip=\fontdimen2\font plus
\BIBentryALTinterwordstretchfactor\fontdimen3\font minus
  \fontdimen4\font\relax}
\providecommand\BIBforeignlanguage[2]{{%
\expandafter\ifx\csname l@#1\endcsname\relax
\typeout{** WARNING: IEEEtran.bst: No hyphenation pattern has been}%
\typeout{** loaded for the language `#1'. Using the pattern for}%
\typeout{** the default language instead.}%
\else
\language=\csname l@#1\endcsname
\fi
#2}}

\bibitem{klein2007parallel}
G.~Klein and D.~Murray, ``Parallel tracking and mapping for small {AR}
  workspaces,'' in \emph{Proc. of the {IEEE} International Symposium on Mixed
  and Augmented Reality}, Nara, Japan, Nov.11--16 2007, pp. 225--234.

\bibitem{engel2014lsd}
J.~Engel, T.~Sch{\"o}ps, and D.~Cremers, ``{LSD-SLAM}: Large-scale direct
  monocular {SLAM},'' in \emph{Proc. of the European Conference on Computer
  Vision}, Zurich, Switzerland, Sept.6--12 2014, pp. 834--849.

\bibitem{mur2015orb}
R.~Mur-Artal, J.~M.~M. Montiel, and J.~D. Tardos, ``{ORB-SLAM}: A versatile and
  accurate monocular slam system,'' \emph{{IEEE} Trans. on Robotics}, vol.~31,
  no.~5, pp. 1147--1163, 2015.

\bibitem{Mourikis07}
A.~I. Mourikis and S.~I. Roumeliotis, ``A multi-state constraint {Kalman}
  filter for vision-aided inertial navigation,'' in \emph{Proc. of the {IEEE}
  International Conference on Robotics and Automation}, Rome, Italy, Apr.
  10--14 2007, pp. 3482--3489.

\bibitem{KFSLAMJ}
S.~Leutenegger, S.~Lynen, M.~Bosse, R.~Siegwart, and P.~Furgale,
  ``Keyframe-based visual-inertial odometry using nonlinear optimization,''
  \emph{International Journal of Robotics Research}, vol.~34, no.~3, pp.
  314--334, March 2015.

\bibitem{qin2018vins}
T.~Qin, P.~Li, and S.~Shen, ``Vins-mono: A robust and versatile monocular
  visual-inertial state estimator,'' \emph{{IEEE} Trans. on Robotics}, vol.~34,
  no.~99, pp. 1--17, 2018.

\bibitem{mur2017visual}
R.~Mur-Artal and J.~D. Tard{\'o}s, ``Visual-inertial monocular {SLAM} with map
  reuse,'' \emph{IEEE Robotics and Automation Letters}, vol.~2, no.~2, pp.
  796--803, 2017.

\bibitem{liu2018ice}
H.~Liu, M.~Chen, G.~Zhang, H.~Bao, and Y.~Bao, ``{ICE-BA}: Incremental,
  consistent and efficient bundle adjustment for visual-inertial {SLAM},'' in
  \emph{Proc. of the {IEEE} Conference on Computer Vision and Pattern
  Recognition}, Salt Lake City, UT, June18-22 2018, pp. 1974--1982.

\bibitem{Triggs00}
B.~Triggs, P.~F. McLauchlan, R.~I. Hartley, and A.~W. Fitzgibbon, ``Bundle
  adjustment - a modern synthesis,'' in \emph{Proc. of the International
  Workshop on Vision Algorithms: Theory and Practice}, ser. Lecture Notes in
  Computer Science, vol. 1883, Corfu, Greece, September 21-22 1999, pp.
  298--372.

\bibitem{Dellaert06squareroot}
F.~Dellaert and M.~Kaess, ``Square root sam: Simultaneous localization and
  mapping via square root information smoothing,'' \emph{International Journal
  of Robotics Research}, vol.~25, no.~12, pp. 1181--1203, December 2006.

\bibitem{isam}
M.~Kaess, A.~Ranganathan, and F.~Dellaert, ``{iSAM}: Incremental smoothing and
  mapping,'' \emph{IEEE Trans. on Robotics}, vol.~24, no.~6, pp. 1365--1378,
  December 2008.

\bibitem{isam2}
M.~Kaess, H.~Johannsson, R.~Roberts, V.~Ila, J.~Leonard, and F.~Dellaert,
  ``{iSAM2}: Incremental smoothing and mapping using the bayes tree,''
  \emph{International Journal of Robotics Research}, vol.~31, no.~2, pp.
  216--235, February 2012.

\bibitem{engel13svo}
J.~Engel, J.~Sturm, and D.~Cremers, ``Semi-dense visual odometry for a
  monocular camera,'' in \emph{Proc. of the {IEEE} International Conference on
  Computer Vision}, Sydney, Australia, Dec.1--8 2013, pp. 1449--1456.

\bibitem{hesch2014consistency}
J.~A. Hesch, D.~G. Kottas, S.~L. Bowman, and S.~I. Roumeliotis, ``Consistency
  analysis and improvement of vision-aided inertial navigation,'' \emph{{IEEE}
  Trans. on Robotics}, vol.~30, no.~1, pp. 158--176, 2014.

\bibitem{SQRT_Invf}
K.~J. Wu, A.~Ahmed, G.~Georgiou, and S.~I. Roumeliotis, ``A square root inverse
  filter for efficient vision-aided inertial navigation on mobile devices,'' in
  \emph{Proc. of Robotics: Science and Systems}, Rome, Italy, July 13 -- 17
  2015.

\bibitem{forster2017svo}
C.~Forster, Z.~Zhang, M.~Gassner, M.~Werlberger, and D.~Scaramuzza, ``{SVO}:
  Semidirect visual odometry for monocular and multicamera systems,''
  \emph{{IEEE} Trans. on Robotics}, vol.~33, no.~2, pp. 249--265, 2017.

\bibitem{bloesch2017iterated}
M.~Bloesch, M.~Burri, S.~Omari, M.~Hutter, and R.~Siegwart, ``Iterated extended
  {Kalman} filter based visual-inertial odometry using direct photometric
  feedback,'' \emph{The International Journal of Robotics Research}, vol.~36,
  no.~10, pp. 1053--1072, 2017.

\bibitem{engel18dso}
J.~Engel, V.~Koltun, and D.~Cremers, ``Direct sparse odometry,'' \emph{{IEEE}
  Trans. on Pattern Analysis and Machine Intelligence}, vol.~40, no.~3, pp.
  611--625, March 2018.

\bibitem{thrun2006graph}
S.~Thrun and M.~Montemerlo, ``The graph {SLAM} algorithm with applications to
  large-scale mapping of urban structures,'' \emph{The International Journal of
  Robotics Research}, vol.~25, no. 5-6, pp. 403--429, 2006.

\bibitem{Konolige08frameslam}
K.~Konolige and M.~Agrawal, ``Frameslam: From bundle adjustment to real-time
  visual mapping,'' \emph{{IEEE} Trans. on Robotics}, vol.~24, no.~5, pp.
  1066--1077, Oct 2008.

\bibitem{nerurkarc}
E.~D. Nerurkar, K.~J. Wu, and S.~I. Roumeliotis, ``C-{KLAM}: Constrained
  keyframe-based localization and mapping for long-term navigation,'' in
  \emph{Proc. of the {IEEE} International Conference on Robotics and
  Automation}, Hong Kong, China, May 31 -- June 7 2014, pp. 3638--3643.

\bibitem{Julier1997}
S.~J. Julier and J.~K. Uhlmann, ``A non-divergent estimation algorithm in the
  presence of unknown correlations,'' in \emph{Proc. of the American Control
  Conference}, vol.~4, Albuquerque, NM, June 4--6 1997, pp. 2369--2373.

\bibitem{Jazwinski1970}
A.~H. Jazwinski, \emph{Stochastic Processes and Filtering Theory}.\hskip 1em
  plus 0.5em minus 0.4em\relax New York, NY: Academic Press, 1970.

\bibitem{Huang2010}
G.~P. Huang, A.~I. Mourikis, and S.~I. Roumeliotis, ``Observability-based rules
  for designing consistent ekf slam estimators,'' \emph{International Journal
  of Robotics Research}, vol.~29, no.~5, pp. 502--528, Apr. 2010.

\bibitem{Mourikis09}
A.~I. Mourikis, N.~Trawny, S.~I. Roumeliotis, A.~E. Johson, A.~Ansar, and
  L.~Matthies, ``Vision-aided inertial navigation for spacecraft entry,
  descent, and landing,'' \emph{{IEEE} Trans. on Robotics}, vol.~25, no.~2, pp.
  264--280, Apr. 2009.

\bibitem{lynen2015get}
S.~Lynen, T.~Sattler, M.~Bosse, J.~Hesch, M.~Pollefeys, and R.~Siegwart, ``Get
  out of my lab: Large-scale, real-time visual-inertial localization,'' in
  \emph{Proc. of Robotics: Science and Systems}, Rome, Italy, July 13-17 2015.

\bibitem{thrun04seif}
S.~Thrun, Y.~Liu, D.~Koller, A.~Y. Ng, Z.~Ghahramani, and H.~Durrant-Whyte,
  ``Simultaneous localization and mapping with sparse extended information
  filters,'' \emph{International Journal of Robotics Research}, vol.~23, no.
  7-8, pp. 693--716, August 2004.

\bibitem{walter07eseif}
M.~R. Walter, R.~M. Eustice, and J.~J. Leonard, ``Exactly sparse extended
  information filters for feature-based {SLAM},'' \emph{International Journal
  of Robotics Research}, vol.~26, no.~4, pp. 335--359, April 2007.

\bibitem{schmidt1966application}
S.~F. Schmidt, ``Application of state-space methods to navigation problems,''
  in \emph{Advances in control systems}.\hskip 1em plus 0.5em minus 0.4em\relax
  Elsevier, 1966, vol.~3, pp. 293--340.

\bibitem{guivant2001optimization}
J.~E. Guivant and E.~M. Nebot, ``Optimization of the simultaneous localization
  and map-building algorithm for real-time implementation,'' \emph{IEEE Trans.
  on Robotics and Automation}, vol.~17, no.~3, pp. 242--257, 2001.

\bibitem{julier2001sparse}
S.~J. Julier, ``A sparse weight {Kalman} filter approach to simultaneous
  localisation and map building,'' in \emph{Proc. of the {IEEE/RSJ}
  International Conference on Intelligent Robots and Systems}, vol.~3, Maui,
  HI, Oct. 29 -- Nov. 3 2001, pp. 1251--1256.

\bibitem{nerurkar2011power}
E.~D. Nerurkar and S.~I. Roumeliotis, ``{Power-SLAM}: A linear-complexity,
  anytime algorithm for {SLAM},'' \emph{The International Journal of Robotics
  Research}, vol.~30, no.~6, pp. 772--788, 2011.

\bibitem{dutoit17consistent}
R.~C. DuToit, J.~A. Hesch, E.~D. Nerurkar, and S.~I. Roumeliotis, ``Consistent
  map-based {3D} localization on mobile devices,'' in \emph{Proc. of the {IEEE}
  International Conference on Robotics and Automation}, 2017, pp. 6253--6260.

\bibitem{inverseschmidt}
K.~J. Wu and S.~I. Roumeliotis, ``Inverse {Schmidt} estimators,''
  \url{http://mars.cs.umn.edu/tr/inverseschmidt.pdf}.

\bibitem{risetechreport}
T.~Ke, K.~J. Wu, and S.~I. Roumeliotis, ``Visual-inertial {SLAM} with inverse
  {Schmidt} estimators,'' \url{http://mars.cs.umn.edu/tr/risetechreport.pdf}.

\bibitem{Burri25012016}
M.~Burri, J.~Nikolic, P.~Gohl, T.~Schneider, J.~Rehder, S.~Omari, M.~W.
  Achtelik, and R.~Siegwart, ``The {EuRoC} micro aerial vehicle datasets,''
  \emph{The International Journal of Robotics Research}, vol.~35, no.~10, pp.
  1157--1163, Sept. 2016.

\bibitem{rublee2011orb}
E.~Rublee, V.~Rabaud, K.~Konolige, and G.~Bradski, ``{ORB}: {A}n efficient
  alternative to {SIFT} or {SURF},'' in \emph{Proc. of the IEEE International
  Conference on Computer Vision}, Barcelona, Spain, Nov. 6--13 2011, pp.
  2564--2571.

\bibitem{nister2006scalable}
D.~Nister and H.~Stewenius, ``Scalable recognition with a vocabulary tree,'' in
  \emph{Proc. of the {IEEE} Conference on Computer Vision and Pattern
  Recognition}, vol.~2, New York, NY, June17--22 2006, pp. 2161--2168.

\bibitem{horn1987closed}
B.~K. Horn, ``Closed-form solution of absolute orientation using unit
  quaternions,'' \emph{JOSA A}, vol.~4, no.~4, pp. 629--642, 1987.

\bibitem{bar2004estimation}
Y.~Bar-Shalom, X.~R. Li, and T.~Kirubarajan, \emph{Estimation with applications
  to tracking and navigation: theory algorithms and software}.\hskip 1em plus
  0.5em minus 0.4em\relax John Wiley {\&} Sons, 2004.

\end{thebibliography}

\end{document}